\documentclass[3p]{elsarticle}

\usepackage{graphicx}
\usepackage{subfigure}

\usepackage{bm}
\usepackage{color}
\usepackage{float}

\usepackage{hyperref}

\journal{ISPRS Journal of Photogrammetry and Remote Sensing}

\begin{document}

\begin{frontmatter}

\title{An Attention-Fused Network for Semantic Segmentation of Very-High-Resolution Remote Sensing Imagery}

\author[mymainaddress,myfifthaddress]{Xuan Yang}
\author[mysecondaryaddress]{Shanshan Li}
\author[mythirdaddress]{Zhengchao Chen}
\author[mymainaddress]{Jocelyn Chanussot}
\author[myforthaddress]{Xiuping Jia}
\author[mymainaddress,myfifthaddress]{Bing Zhang\corref{mycorrespondingauthor}}
\cortext[mycorrespondingauthor]{Corresponding author at: Aerospace Information Research Institute, Chinese Academy of Sciences, Beijing 100094, China}
\ead{zb@radi.ac.cn}
\author[mythirdaddress]{Baipeng Li}
\author[mymainaddress,myfifthaddress]{Pan Chen}

\address[mymainaddress]{Key Laboratory of Digital Earth Science, 
Aerospace Information Research Institute, Chinese Academy of Sciences, Beijing 100094, China}
\address[mysecondaryaddress]{China Remote Sensing Satellite Ground Station, 
Aerospace Information Research Institute, Chinese Academy of Sciences, Beijing 100094, China}
\address[mythirdaddress]{Airborne Remote Sensing Center, 
Aerospace Information Research Institute, Chinese Academy of Sciences, Beijing 100094, China}
\address[myforthaddress]{School of Engineering and Information Technology, The University of New South Wales, 
Australian Defence Force Academy, Canberra, A.C.T. 2612, Australia}
\address[myfifthaddress]{University of Chinese Academy of Sciences, Beijing 100049, China}

\begin{abstract}
\textcolor{blue}{This is a preprint version of a paper accepted by ISPRS Journal of Photogrammetry and Remote Sensing.}

\noindent
Semantic segmentation is an essential part of deep learning. In recent years, with the development of remote sensing big data, semantic segmentation has been increasingly used in remote sensing. Deep convolutional neural networks (DCNNs) face the challenge of feature fusion: very-high-resolution remote sensing image multisource data fusion can increase the network’s learnable information, which is conducive to correctly classifying target objects by DCNNs; simultaneously, the fusion of high-level abstract features and low-level spatial features can improve the classification accuracy at the border between target objects. In this paper, we propose a multipath encoder structure to extract features of multipath inputs, a multipath attention-fused block module to fuse multipath features, and a refinement attention-fused block module to fuse high-level abstract features and low-level spatial features. Furthermore, we propose a novel convolutional neural network architecture, named attention-fused network (AFNet). Based on our AFNet, we achieve state-of-the-art performance with an overall accuracy of 91.7\% and a mean F1 score of 90.96\% on the ISPRS Vaihingen 2D dataset and an overall accuracy of 92.1\% and a mean F1 score of 93.44\% on the ISPRS Potsdam 2D dataset. 
\end{abstract}


\end{frontmatter}


\section{Introduction}
\label{sec:1}

In recent years, with the rapid development of remote sensing technology, the amount of remote sensing data that has been obtained has grown significantly \cite{ma2015remote}. Remotely sensed big data have 4Vs characteristics, which represent volume, variety, velocity, and veracity \cite{zhang2018remotely,zhang2019remotely}. We can exploit rich and important information from remotely sensed big data. With the improvement in sensor technology, the spatial resolution of remote sensing images is increasing. In high spatial resolution images, the spatial texture details of target objects are preserved \cite{trias2008using}. We can use spatial texture information to identify, classify, and even extract accurate contours to exploit rich geological spatial information contained in images. The higher the spatial resolution is, the larger the volume of data and the richer the information it contains \cite{carleer2005assessment}. The high-resolution remote sensing imagery's spatial resolution can reach the meter or decimeter level, while the very-high-resolution imagery can reach the centimeter level. In very-high-resolution images, each target object has rich details. We can distinguish and identify different target objects based on these detailed features. Some target objects must be accurately identified by very-high-resolution images \cite{benediktsson2012very}. In low- and medium-resolution images, some similar target objects are easily confused and difficult to distinguish from each other due to the loss of a large amount of texture information. Therefore, very-high-resolution images can be more accurately used in target object recognition and classification and have an advantage over low- and medium-resolution images. 

In recent years, deep learning has been developed by leaps and bounds in the field of computer vision \cite{lecun2015deep}. Deep learning is a data-driven technology \cite{reichstein2019deep}. With the development of big data, deep learning has significant advantages \cite{chen2014big}. Deep learning for image analysis is based on deep convolutional neural networks (DCNNs), building complex spatial texture expression models and exploiting content information in images. Deep learning is widely used in applications, such as scene classification \cite{lecun1998gradient,krizhevsky2012imagenet,szegedy2015going,simonyan2014very}, object detection \cite{girshick2014rich,girshick2015fast,ren2015faster,liu2016ssd}, and semantic segmentation \cite{long2015fully,badrinarayanan2017segnet,ronneberger2015u,noh2015learning}. Among these applications, semantic segmentation is the classification of each pixel in a picture, which is a kind of pixel-level image classification. Since all pixels are classified, the contours of different types of target objects can be accurately extracted. The positions, shapes, and spatial distribution of the target objects are more accurate. 

In the field of remote sensing, the typical applications of semantic segmentation are land-use mapping \cite{castelluccio2015land,cheng2015effective,hu2013automated}, land-cover mapping \cite{friedl1997decision,running1995remote,townshend1991global}, building extraction \cite{lefevre2007automatic,vu2009multi}, waterbody extraction \cite{zhaohui2003water,shen2010water}, and so on. Semantic segmentation based on traditional remote sensing methods requires the artificial design of corresponding feature extractors according to the characteristics of different target objects. The artificially designed feature extractors have high professional knowledge requirements \cite{ball2017comprehensive}, cannot adapt to complex application scenarios and have limited generalization capabilities. Deep learning-based semantic segmentation can effectively overcome the limitations of traditional remote sensing methods \cite{zhang2016deep}. This method can extract rich features and has strong robustness. DCNN learns the feature information of different target objects by itself, thereby achieving pixel-level image classification, and {the method} has a strong generalization ability. 

However, there are also some difficulties in the application of deep learning in the field of remote sensing, and these difficulties are outlined as follows: 

\begin{itemize}
\item Images in the field of computer vision are generally RGB three-channel images. However, remote sensing images are composed of multiband data. There are also some other types of remote sensing data, such as the normalized difference vegetation index (NDVI) and digital surface model (DSM). These data are not obtained by optical sensors and have different characteristics from ordinary optical images. The most popular DCNNs work with three-channel RGB optical images. Although those DCNNs can work with single-channel or multichannel images, it is not appropriate if we simply stack the optical data and the other structural data. It is harder to train a network by using one encoder to extract multisource data features than by using individual encoders to learn the individual modalities. Fusing the separate features in the decoder will simplify the training objective. Current fusion methods for the features extracted from multisource data rely on summing the feature maps\cite{audebert2018beyond,audebert2016semantic} or concatenating individual feature maps\cite{sherrah2016fully,marmanis2018classification}. The effective fusion of the features remains an open research direction. 
\item The DCNN is a stack of many convolutional layers and pooling layers. The convolutional layer is used to extract features, and the pooling layer is used to aggregate features. The deeper the network is, the more abstract the extracted information. However, in the pooling layer, a significant amount of spatial information is lost. The shallow part of the network cannot adequately extract abstract information, but the spatial information is kept intact. Semantic segmentation must be able to both extract abstract information and retain more accurate position information to achieve correct pixel-level image classification. The scenes of remote sensing images are very complicated. The effective fusion of low-level spatial features and high-level abstract features is a problem that needs to be further optimized. 
\end{itemize}

In summary, these difficulties include two types of feature fusion: 1) multipath feature fusion extracted from multisource data and 2) multilevel feature fusion for high-level abstract features and low-level spatial features. However, mainstream DCNNs cannot yet efficiently and effectively deal with the problems of feature fusion. In this paper, we propose a novel attention-fused network (AFNet) architecture, including the multipath attention-fused block (MAFB) module and refinement attention-fused block (RAFB) module, which perform well in solving the problems of “multipath feature fusion” and “multilevel feature fusion”. 

The MAFB module is designed to solve the difficulty of “multipath feature fusion”. In the task of semantic segmentation for target objects, data from different sources may play a key role. Therefore, to ensure that multipath features extracted from different inputs are treated equally, we use a symmetric structure to feed these features into MAFB. To suppress the interference of useless feature information on the classification results, we introduce an attention structure. We use a channel attention \cite{hu2018squeeze} module to calculate the feature weights in the channel dimension and obtain the key channel features. We use a spatial attention \cite{woo2018cbam} module to calculate the feature weights in the spatial dimension to obtain the key spatial features. The fusion of these two key features completes the selection and fusion of the multipath features. 

The RAFB module is designed to solve the difficulty of “multilevel feature fusion”. We use a channel attention module to calculate the feature weights in the channel dimension from the high-level abstract features and then use the feature weights to select the useful low-level spatial features to improve the abstract expression ability of the low-level spatial features. We use a spatial attention module to calculate the feature weights in the spatial dimension from the low-level spatial features and then use the feature weights to refine the spatial details of the high-level abstract features. Finally, we fuse these two refined features and obtain the fused multilevel features. 

\begin{figure}[ht]
\centering
\includegraphics[width=1.0\linewidth]{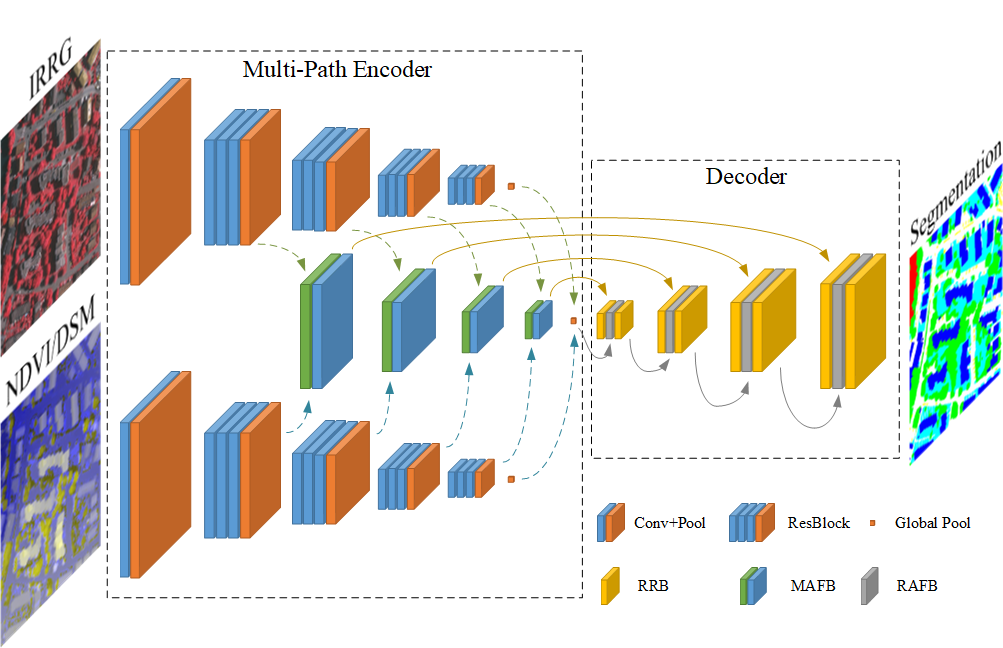}
\caption{Overview of the Attention-Fused Network (AFNet) architecture.}
\label{fig:afnet3d}
\end{figure}

In summary, the contributions of this paper are described as follows: 

\begin{itemize}
\item Inspired by the channel attention structure and the spatial attention structure, we design a variant spatial attention module. The variant spatial attention module is designed to calculate the feature weights in the spatial dimension and extract useful key spatial features. 
\item We design a multipath encoder (MPE) structure to simultaneously extract the abstract features and the spatial features from the different data input sources. We rethink the method of feature fusion in the DCNN and design a multipath attention-fused block (MAFB) module to fuse the multipath features from the MPE structure. 
\item We design a refinement attention-fused block (RAFB) module to fuse low-level spatial features and high-level abstract features. According to the characteristics of different level features, the RAFB module makes full use of the advantages of those features. 
\item By integrating the MPE structure with the MAFB module and the RAFB module, we propose an attention-fused network (AFNet) to simultaneously address the “multipath feature fusion” and “multilevel feature fusion” issues. An overview of the AFNet architecture is shown in Figure \ref{fig:afnet3d}. Our proposed AFNet achieves state-of-the-art performances on the ISPRS Vaihingen 2D dataset and the ISPRS Potsdam 2D dataset \cite{webisprscontest}. 
\end{itemize}

The remainder of this paper is organized as follows: Section \ref{sec:2} presents the related work. In Section \ref{sec:3}, we introduce our proposed methodology about the multipath V-shape network (MPVN), MAFB, RAFB, and AFNet architecture. Section \ref{sec:4} experimentally validates AFNet on the ISPRS Vaihingen 2D dataset and ISPRS Potsdam 2D dataset. In Section \ref{sec:5}, we discuss the impact of training parameters on AFNet. Section \ref{sec:6} presents the conclusion of this paper. 

\section{Related Work}
\label{sec:2}

Several popular CNNs developed in recent years, such as AlexNet \cite{krizhevsky2012imagenet}, VGGNet \cite{simonyan2014very}, GoogLeNet \cite{szegedy2015going}, and ResNet \cite{he2016deep}, have been used in scene classification. FCN \cite{long2015fully} is the first fully convolutional neural network that is designed for semantic segmentation. FCN uses skip connections to refine feature maps and upsample the output feature maps to the size of the input origin data. However, the abstraction ability of FCN for the high-level features is not enough to consider useful global context information. The precise boundary cannot be accurately restored by eight times upsampling at the end of the FCN. The FCN is outperformed by other state-of-the-art methods. 

Subsequently, two styles of structure appear in the semantic segmentation network. One is the backbone style, such as PSPNet \cite{zhao2017pyramid} and DeepLabV3 \cite{chen2017rethinking}. This type of network uses dilated convolution instead of ordinary convolution in the encoder. This network type removes some pooling layers and reduces the downsampling degree in feature maps. These networks use the pyramid pooling module \cite{zhao2017pyramid} or atrous spatial pyramid pooling (ASPP) \cite{chen2017rethinking} module to extract and fuse feature information of different scales and receptive fields. However, dilated convolution makes the training and inference processes time-consuming and memory intensive. Additionally, the precise boundary cannot be accurately restored by eight times upsampling at the decoder in these two methods. The other type is the encoder-decoder style, such as UNet \cite{ronneberger2015u} and SegNet \cite{badrinarayanan2017segnet}. The decoder in this type of network gradually upsamples, and the spatial information lost due to pooling is gradually restored. During upsampling, the feature maps in the decoder are fused by the skip connection with the feature maps of the corresponding stage in the encoder of UNet. SegNet uses the saved pool indices to restore the reduced spatial information. However, the restoration and upsampling operations are too simple to refine the abstract features, leading to some contradictory results. 

RefineNet \cite{lin2017refinenet} is also an encoder-decoder style network. This network uses a multipath refinement network to refine the feature but ignores the global context feature. ParseNet \cite{liu2015parsenet} first uses a global pooling module to extract global features. PSPNet and DeepLabV3 also adopt a global pooling module in their networks. We use the global context feature in this paper. However, these methods only use concatenation operation to combine the features of different receptive fields and ignore their diverse feature representations. A channel attention structure is used in SENet \cite{hu2018squeeze}. This structure allows the neural network to recognize the critical channels of the feature map and select the most suitable channels by itself. However, SENet only applies the attention structure to the channel dimension of the feature map and not the spatial dimension. In fact, spatial attention is also important for semantic segmentation. DANet \cite{fu2019dual} uses both channel attention and spatial attention to refine the feature. The difference is that the attention structure in DANet is based on the self-attention structure \cite{yuan2018ocnet}. The two attention branches independently extract single-path features and perform simple fusion through addition. In this paper, both the channel attention structure and the spatial attention structure are used to constrain and guide each other to fuse the multipath features and different level features. 

The discriminative feature network (DFN) \cite{yu2018learning} is a high-performance semantic segmentation network that achieves state-of-the-art performance on the Cityscapes dataset \cite{cordts2016cityscapes}. The DFN uses channel attention block (CAB) and refinement residual block (RRB) to solve the aforementioned problems. This network is also an encoder-decoder style network. The encoder is ResNet, and a global pooling module is used in DFN. The decoder includes the residual connection module and channel attention module. The DFN uses the channel attention module to fuse abstract features and spatial features. However, the DFN does not consider multipath inputs and does not use spatial feature weights to refine the high-level features. The DFN limits the full utilization of data and the feature fusion performance. 

Thanks to the development of computer vision, the semantic segmentation of very-high-resolution remote sensing imagery can be further developed based on existing DCNNs. DST\_2 \cite{sherrah2016fully} removes the pooling layer from the FCN to retain accurate spatial features. However, similar to the original FCN, eight times upsampling makes the boundaries of target objects less precise. UFMG\_4 \cite{nogueira2019dynamic} uses dilated convolution instead of ordinary convolution to achieve a similar effect. However, similar to PSPNet and DeepLabV3, dilated convolution makes the network's training very slow. This method is based on UNet so that the features from different stages are directly summed without any careful selection. ONE\_7 \cite{audebert2016semantic} uses two encoder branches to learn multisource features to help improve the accuracy of semantic segmentation. However, this method does not use the global context feature to distinguish among various categories. Additionally, this method does not use the attention structure for the selection of useful features. 

\begin{figure}[ht]
\centering
\includegraphics[width=1.0\linewidth]{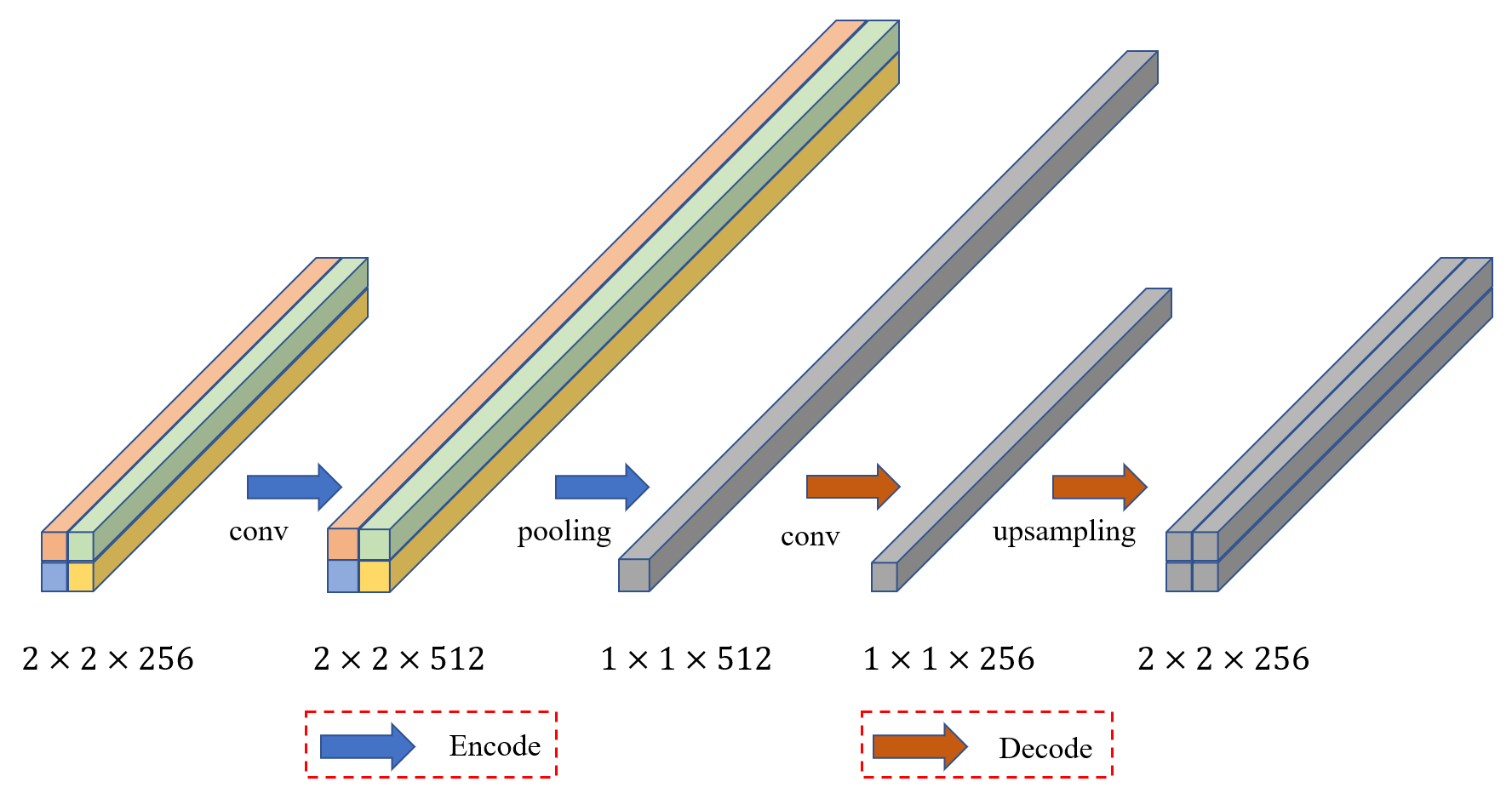}
\caption{The position information is severely lost after encoding. The original position information cannot be restored even if we upsample the encoded features.}
\label{fig:down_upsample}
\end{figure}

\begin{figure}[H]
\centering
\subfigure[]{
    \includegraphics[width=0.73\linewidth]{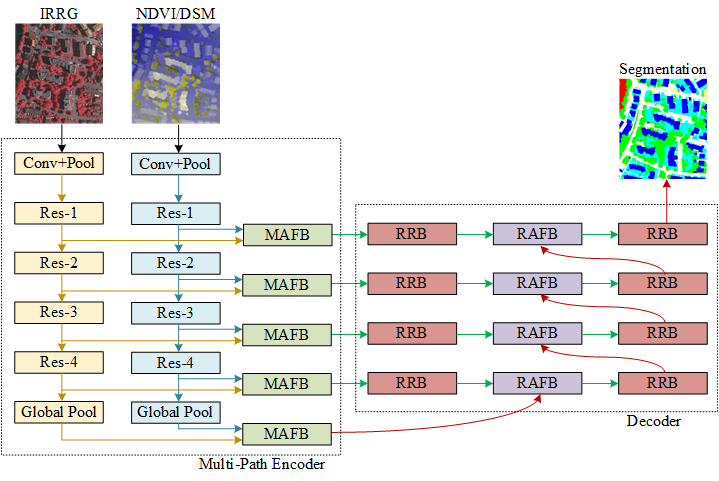}
    \label{fig:afnet:main}
}
\vfill
\subfigure[]{
    \includegraphics[width=0.73\linewidth]{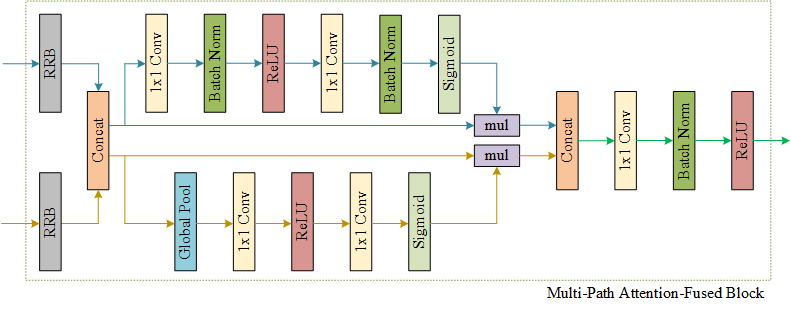}
    \label{fig:afnet:mafb}
}
\vfill
\subfigure[]{
    \includegraphics[width=0.53\linewidth]{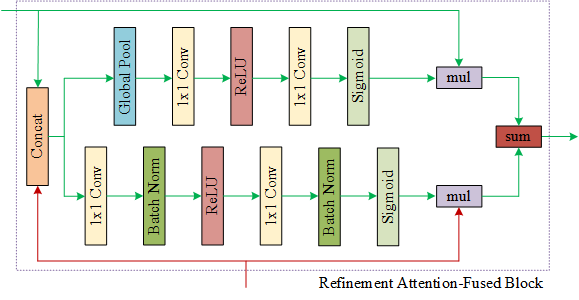}
    \label{fig:afnet:rafb}
}
\caption{Details of the attention-fused network (AFNet) architecture. (a) Main structure of AFNet. The “RRB” block represents the refinement residual block module in Figure \ref{fig:dfn:rrb}. (b) Details of the multipath attention-fused block (MAFB) module. The “mul” block represents dot multiplication of two features. (c) Details of the eefinement attention-fused block (RAFB) module. The “mul” block represents dot multiplication of two features, and the “sum” block represents the simple summation of two features.}
\label{fig:afnet}
\end{figure}

\section{Methodology}
\label{sec:3}

DCNNs rely on encoders to extract features. On the one hand, the feature map is downsampled multiple times in the encoder to save the cost of hardware resources, and on the other hand, it is downsampled multiple times to aggregate feature information and increase the receptive field of the CNN. As the encoder gradually deepens, the feature information becomes increasingly abstract, and the feature expression ability becomes increasingly stronger. However, as shown in Figure \ref{fig:down_upsample}, a significant amount of the precise position information is lost after encoding. When high-level features are upsampled back to the original size, the original position information cannot be accurately restored. In deep learning for semantic segmentation, we need to restore the size of the feature map to the same size as the original image to achieve pixel-level classification. This process is implemented by the decoder. The decoder fuses low-level spatial features with high-level abstract features. Therefore, the neural network not only has a good classification performance but also retains more accurate spatial information. 

The imaging platform of remote sensing images is generally located at high altitudes, making some target objects appear small on the remote sensing images. This platform makes it easy for the encoder to skip small target objects. Therefore, the best network architecture to solve the small target object problem is the encoder-decoder style architecture, which can simultaneously consider both the abstraction ability and the position information. In this paper, the DFN is used as the baseline network. According to the characteristics of remote sensing images, we proposed a novel DCNN called attention-fused network (AFNet). The details of the AFNet architecture are shown in Figure \ref{fig:afnet}. For the feature extraction of multipath inputs, a multipath encoder (MPE) is designed for AFNet. For the feature fusion of multipath features, a multipath attention-fused block (MAFB) is designed for AFNet. For the fusion of abstract features and spatial information, inspired by the CAB in the DFN, a refinement attention-fused block (RAFB) is designed for AFNet. 

\subsection{Baseline Network}

\begin{figure}[H]
\centering
\subfigure[]{
    \includegraphics[width=0.6\linewidth]{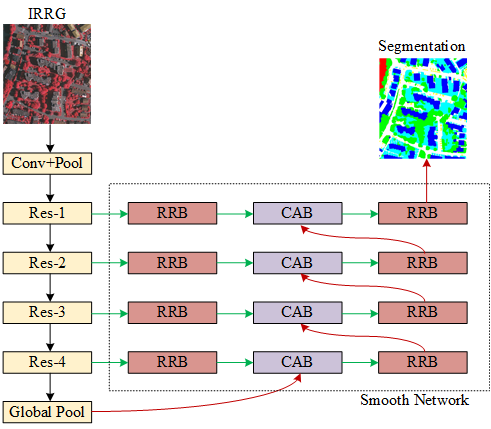}
    \label{fig:dfn:main}
}
\vfill
\subfigure[]{
    \includegraphics[width=0.5\linewidth]{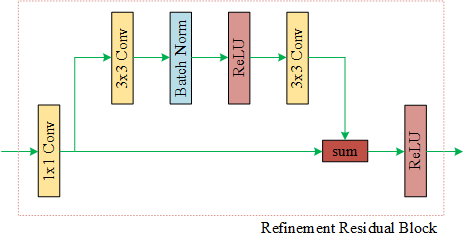}
    \label{fig:dfn:rrb}
}
\vfill
\subfigure[]{
    \includegraphics[width=0.7\linewidth]{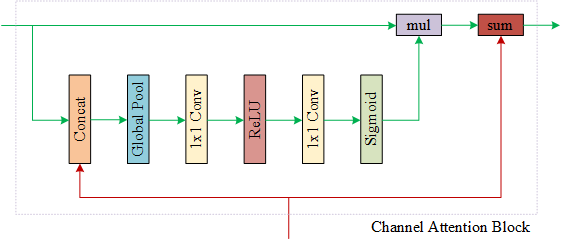}
    \label{fig:dfn:cab}
}
\caption{Overview of the discriminative feature network (DFN) architecture. (a) Main structure of the DFN without the border network. (b) Details of the refinement residual block (RRB) module. (c) Details of the channel attention block (CAB) module.}
\label{fig:dfn}
\end{figure}

We use the DFN as the baseline network architecture, which is a V-shape CNN and belongs to the encoder-decoder style network. The encoder consists of ResNet and a global pooling module. ResNet in the DFN removes the last pooling layer and fully connected layer in the original ResNet. Finally, the encoder outputs a 2048-dimensional (ResNet-50/101/152) or 512-dimensional (ResNet-18/34) feature map that is 1/32 the size of the original image. Then, to obtain global features, the DFN adds a global pooling module after ResNet to obtain a feature vector with a length of 2048 or 512 dimensions. A network structure called the smooth network is designed as the decoder of the DFN. The smooth network contains two types of network modules. One type is the CAB. Firstly, the CAB fuses the high-level abstract features and low-level spatial features. Then, the CAB uses the attention structure to learn feature weights. Finally, the CAB fuses the weighted low-level spatial features and the original high-level abstract features. In other words, the network uses high-level abstract features to guide the selection of effective low-level spatial features. The module solves the problem of insufficient abstraction of low-level features and the problem of severe loss of high-level feature spatial information. The other type is the RRB. The RRB uses the residual structure to refine the abstract features and improve the abstract expression ability of the features. At the same time, the RRB avoids the problem of gradient disappearance caused by too many network layers. In the DFN, there is another branch network structure called the border network. The border network learns the boundary feature information of the original image. The network uses the boundary feature information to constrain the features in the encoder of the DFN to improve the accuracy of object boundaries. On the ISPRS Vaihingen 2D dataset and ISPRS Potsdam 2D dataset, because the boundaries of objects are ignored during the evaluation, the border network is not helpful for the segmentation results. Therefore, we remove the border network in the baseline DFN. The baseline DFN architecture without the border network is shown in Figure \ref{fig:dfn}. 

\begin{figure}[H]
\centering
\includegraphics[width=1.0\linewidth]{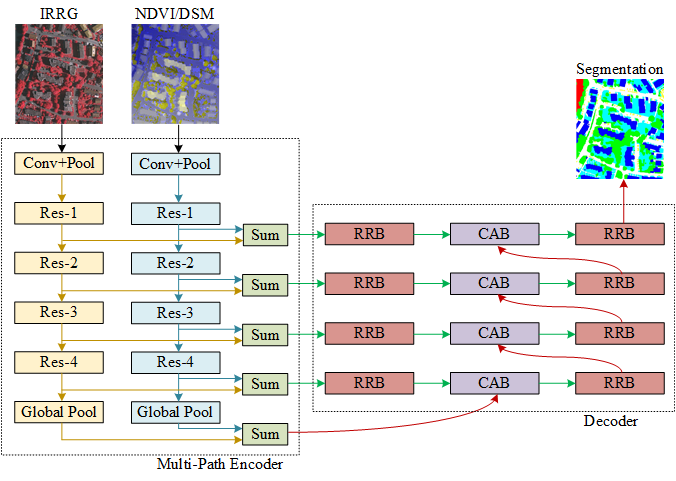}
\caption{Overview of the multipath V-shape network (MPVN) architecture. The “Sum” block represents the simple summation of the multipath features. The “RRB” block represents the refinement residual block module in Figure \ref{fig:dfn:rrb}. The “CAB” block represents the channel attention block module in Figure \ref{fig:dfn:cab}.}
\label{fig:mpvn}
\end{figure}

\subsection{Multipath Encoder}

The baseline DFN is designed for the Cityscapes dataset. The input of the encoder is RGB three-channel pictures. In the field of remote sensing, there are often auxiliary data, such as DSM, or simple feature data, such as NDVI. The DFN encoder cannot simultaneously extract features from image data and auxiliary data. We propose a multipath encoder (MPE) that replaces the original encoder in the DFN. the MPE has two branches, the main branch and the auxiliary branch, for extracting the features of the image data and the features of the auxiliary data, respectively. The DSM data are used to describe the surface elevation information of the corresponding localization of each pixel in the image, which is similar to depth information in the field of computer vision. Compared with the complex color characteristic structure in the image data, the characteristic structure of the DSM data is relatively simple. Index features, such as NDVI, can be regarded as feature maps after simple encoding. Therefore, the auxiliary data can use a relatively simple encoder to extract features. A simple encoder can avoid network overfitting, save computing resources and increase computing speed. 

After the first convolution layer, ResNet outputs a 64-dimensional feature map. We can combine the image data and the auxiliary data into a multichannel image. Then, we use the encoder in the DFN to extract the features of the multichannel image. However, the feature expression ability for multichannel images is limited in the DFN. Arguably, we more often extract multipath features independently and then fuse them. The auxiliary data can be regarded as a simple encoding feature that is totally different from the multichannel image. However, the MPE uses two branches to extract different types of features. The MPE can improve the feature expression ability and make feature expression clearer for different types of data. Then two features are fused by simply summation, and the decoder is used to gradually restore the feature map size to the original image size. Therefore, we propose a multipath V-shape network (MPVN) based on the DFN. The MPVN architecture is shown in Figure \ref{fig:mpvn}. 

\subsection{Multipath Attention-Fused Block}

The MPVN fuses two types of features by using a common fusion method but ignores the importance of the different types of features. However, different target objects have various sensitivities to varying types of features. Color and texture are the main features that indicate target objects. NDVI can be used to distinguish vegetation and some confusing features. DSM plays a significant role for some target objects closely related to their height above the ground surface. Therefore, different types of features extracted by the MPE need to be assigned appropriate weights according to different target objects. The MPE with a better feature fusion method can improve the performance of the entire network. We propose a new feature-fusion module to replace the simply summed module of the MPVN. 

\begin{figure}[ht]
\centering
\includegraphics[width=1.0\linewidth]{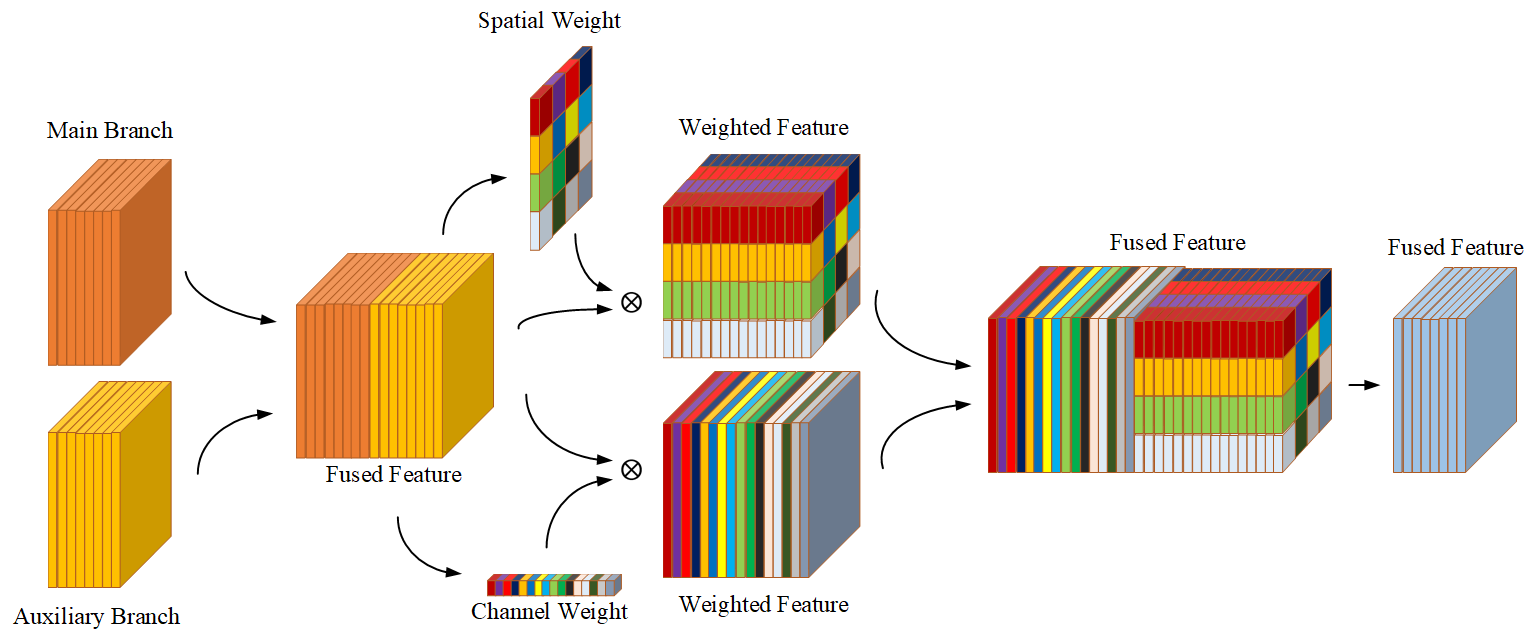}
\caption{Schematic diagram of the multipath attention-fused block (MAFB) module. The orange block represents the feature of the main branch, and the yellow block represents the auxiliary branch. We concatenate two branches of features to compute the spatial weight vector and channel weight vector. We use those two vectors to weight the fused feature map. Finally, we fuse the weighted feature maps. The color in the weight vector represents the weight value of the spatial dimension and the channel dimension.}
\label{fig:mafb3d}
\end{figure}

The CAB module allows the MPVN to learn the attention weights by itself on the channel dimension and uses this weight to mine the effective feature information required. However, the MPVN does not pay attention to spatial weights. The attention weights in the spatial dimension are also important. The spatial texture features required by different types of target objects are different. We introduce the idea of spatial attention to the MPVN. These two attention modules allow the network to simultaneously learn the weights of the channel and spatial dimensions. 

First, we express a convolution layer ${\rm W}^{n}(x)$ as follows: 

\begin{equation}
{\rm W}^{n}(x)=\mathbf{W}^{n\times n}\odot x+\mathbf{b}
\label{eq:conv}
\end{equation}

\noindent
where $\odot$ represents the convolution operator, $\mathbf{W}^{n\times n}$ represents the $n\times n$ convolutional kernel, $\mathbf{b}$ represents the vector of bias, and $x$ represents the input data. 

The channel attention (CA) module in the DFN can be expressed as follows: 

\begin{equation}
f_{\rm CA}(x)=f_{\rm sigmoid}({\rm W}_2^1(f_{\rm ReLU}({\rm W}_1^1(f_{\rm AvgPool}^1(x)))))
\label{eq:ca}
\end{equation}

\noindent
where $f_{\rm AvgPool}^1$ represents the function of global average pooling, $f_{\rm Sigmoid}$ represents the sigmoid function, $f_{\rm ReLU}$ represents the activation function of the rectified linear unit, ${\rm W}_1^1$ and ${\rm W}_2^1$ represents the first and second $1\times 1$ convolution layer, respectively, and $x$ represents the input data. 

Inspired by the CA module, we design a spatial attention (SA) module with a similar structure. The idea of the SA module is given by 

\begin{equation}
f_{\rm SA}(x)=f_{\rm sigmoid}({\rm W}_2^1(f_{\rm ReLU}({\rm W}_1^1(x))))
\label{eq:sa}
\end{equation}

\noindent
where $f_{\rm sigmoid}$ represents the sigmoid function, $f_{\rm ReLU}$ represents the activation function of the rectified linear unit, ${\rm W}_1^1$ and ${\rm W}_2^1$ represents the first and second $1\times 1$ convolution layer, respectively, and $x$ represents the input data. 

Combining the CA module and SA module, we propose the multipath attention-fused block (MAFB), which uses the idea of the attention structure, as shown in Figure \ref{fig:mafb3d}. MAFB allows the network to learn the feature weights of different target objects and improves the effectiveness of multisource feature fusion. More details of MAFB are shown in Figure \ref{fig:afnet:mafb}, where the features of the two branches are connected to the RRB to ensure that the feature dimensions are consistent. Then, the two features are concatenated to obtain the combined feature. Next, a SA module and a CA module simultaneously connect to the combined features. Equation \ref{eq:sa} and Equation \ref{eq:ca} can learn the SA weight and CA weight, respectively. The two attention weights are weighted to the combined feature. We obtain two new weighted features. Then, we concatenate the two new weighted features and reduce the dimension of the feature map to 512 by the convolution operation. The MAFB module allows the network to learn the effective information in the image data and the auxiliary data by itself. The module can suppress useless information and interference information. 

\begin{figure}[ht]
\centering
\includegraphics[width=1.0\linewidth]{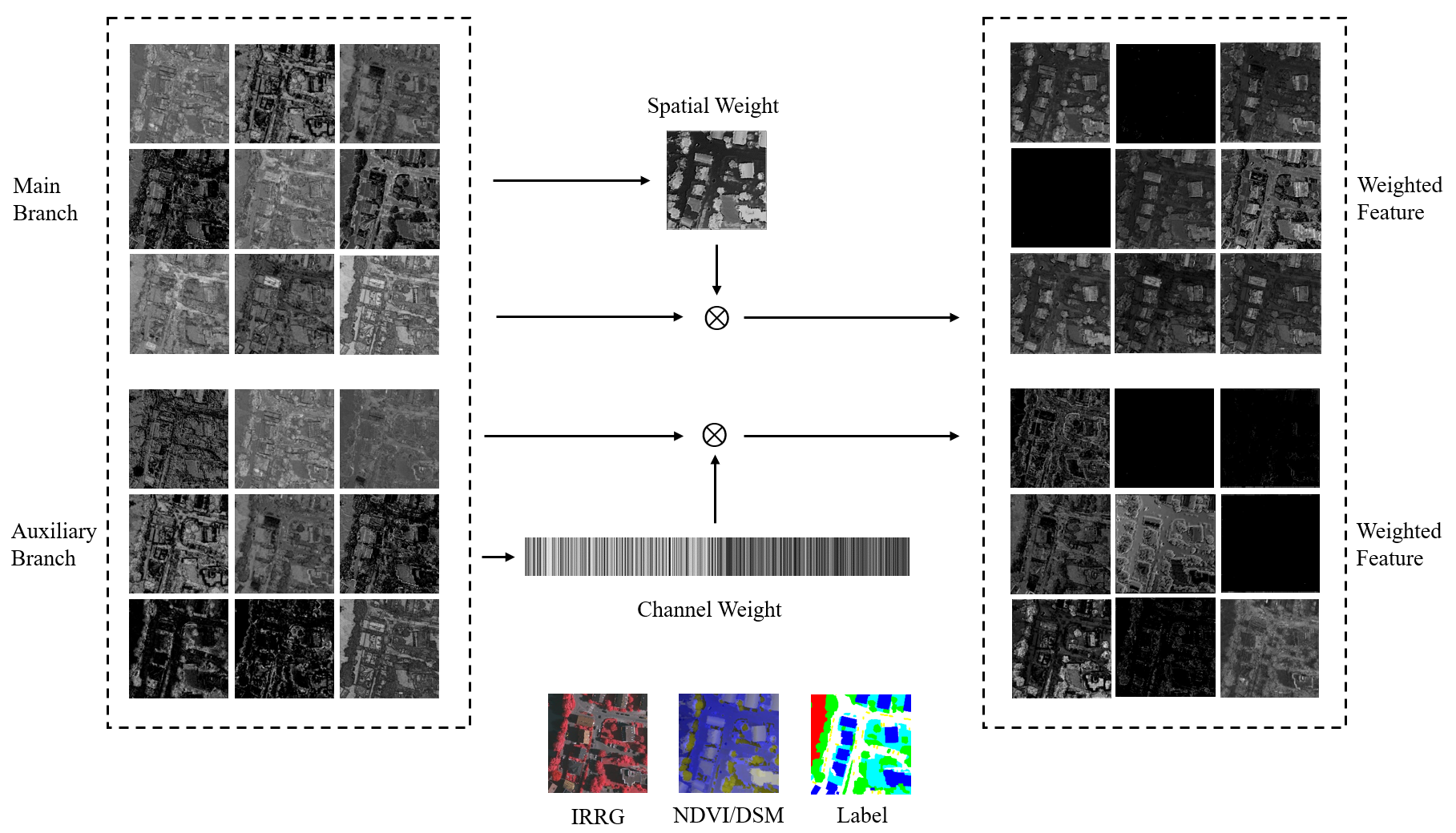}
\caption{Visualization of the feature maps in the MAFB module.}
\label{fig:mafbvis}
\end{figure}

We denote the concatenation operation as follow: 

\begin{equation}
x_{\rm concat}=x_1\oplus x_2
\label{eq:concat}
\end{equation}

\noindent
where $\oplus$ represents the concatenation operator and $x_1$ and $x_2$ represents the features of the two branches. 

The MAFB module can be denoted as follow: 

\begin{equation}
y_{\rm MAFB}={\rm W}_3^1((f_{\rm SA}(x_{\rm concat})\otimes x_{\rm concat})\oplus (f_{\rm CA}(x_{\rm concat})\otimes x_{\rm concat}))
\label{eq:mafb}
\end{equation}

\noindent
where $\oplus$ represents the concatenation operator, $\otimes$ represents the dot multiply operator, $f_{\rm CA}$ represents the CA module mentioned in Equation \ref{eq:ca}, $f_{\rm SA}$ represents the SA module mentioned in Equation \ref{eq:sa}, ${\rm W}_3^1$ represents the last $1\times 1$ convolution layer used for reducing the dimension of the feature map, and $x_{\rm concat}$ represents the combined feature. 

As shown in Figure \ref{fig:mafbvis}, the MAFB module simultaneously learns the SA weight and the CA weight from the simply fused features. The brighter area in the spatial weight map is the most important spatial contextual information learned by the network. The feature in the brighter area is strengthened, and other unimportant spatial features are suppressed. The CA weight allows the network to select the useful feature maps for classification from different branches. The network pays more attention to the critical information during feature fusion by the MAFB module. 

\subsection{Refinement Attention-Fused Block}

There is a trade-off in the network architecture of the pixel-level classification of semantic segmentation. The high-level features have a high degree of abstraction, but the accuracy of the spatial information is low. The low-level features have more accurate spatial information, but the abstraction ability is insufficient. All the encoder-decoder style networks attempt to achieve a balance by fusing low-level features and high-level features. We propose a module to fuse the two levels of features. 

\begin{figure}[ht]
\centering
\includegraphics[width=1.0\linewidth]{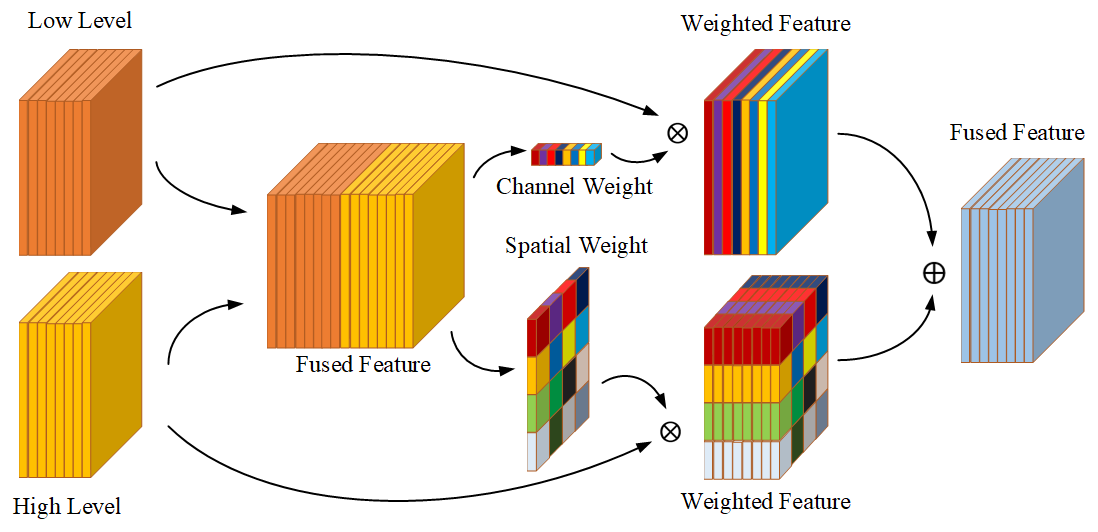}
\caption{Schematic diagram of the refinement attention-fused block (RAFB) module. The orange block represents the low-level feature, and the yellow block represents the high-level feature. We concatenate two levels of features to compute the spatial weight vector and channel weight vector. We use the spatial weight vector to weight the low-level feature map and use the channel weight vector to weight the high-level feature map. Finally, we sum up the two weighted feature maps. The color in the weight vector represents the weight value of the spatial dimension and the channel dimension.} 
\label{fig:rafb3d}
\end{figure}

The MPVN only uses CA weights to employ high-level abstract features to guide the selection of low-level spatial features. In turn, we can infer that low-level spatial features can also guide the selection of high-level abstract features. Therefore, we propose the refinement attention-fused block (RAFB), which introduces the idea of SA, as shown in Figure \ref{fig:rafb3d}. The RAFB module allows the network to learn the weights in the spatial dimension by itself and improves the spatial localization accuracy of high-level abstract features. More details regarding the RAFB are shown in Figure \ref{fig:afnet:rafb}, where the low-level spatial features and the high-level abstract features are concatenated to obtain the combined feature. Next, a SA module and a CA module are simultaneously connected to the combined feature. Equation \ref{eq:sa} and Equation \ref{eq:ca} can be used to learn the SA weight and CA weight, respectively. Unlike the MAFB module, in the RAFB, the SA weight is assigned to high-level abstract features, and the CA weight is assigned to low-level spatial features. Then, we add the two weighted features to obtain the fused feature. The RAFB module takes advantage of the high-level abstract features and low-level spatial features and ultimately improves the accuracy of pixel-level classification in the semantic segmentation . 

The RAFB module can be expressed as follows: 

\begin{equation}
y_{\rm RAFB}=f_{\rm CA}(x_1\oplus x_2)\otimes x_1 + f_{\rm SA}(x_1\oplus x_2)\otimes x_2
\label{eq:rafb}
\end{equation}

\noindent
where $\oplus$ represents the concatenation operator, $\otimes$ represents the dot multiply operator, $f_{\rm CA}$ represents the CA module mentioned in Equation \ref{eq:ca}, $f_{\rm SA}$ represents the SA module mentioned in Equation \ref{eq:sa}, $x_1$ represents the low-level spatial features, and $x_2$ represents the high-level abstract features. 

\begin{figure}[ht]
\centering
\includegraphics[width=1.0\linewidth]{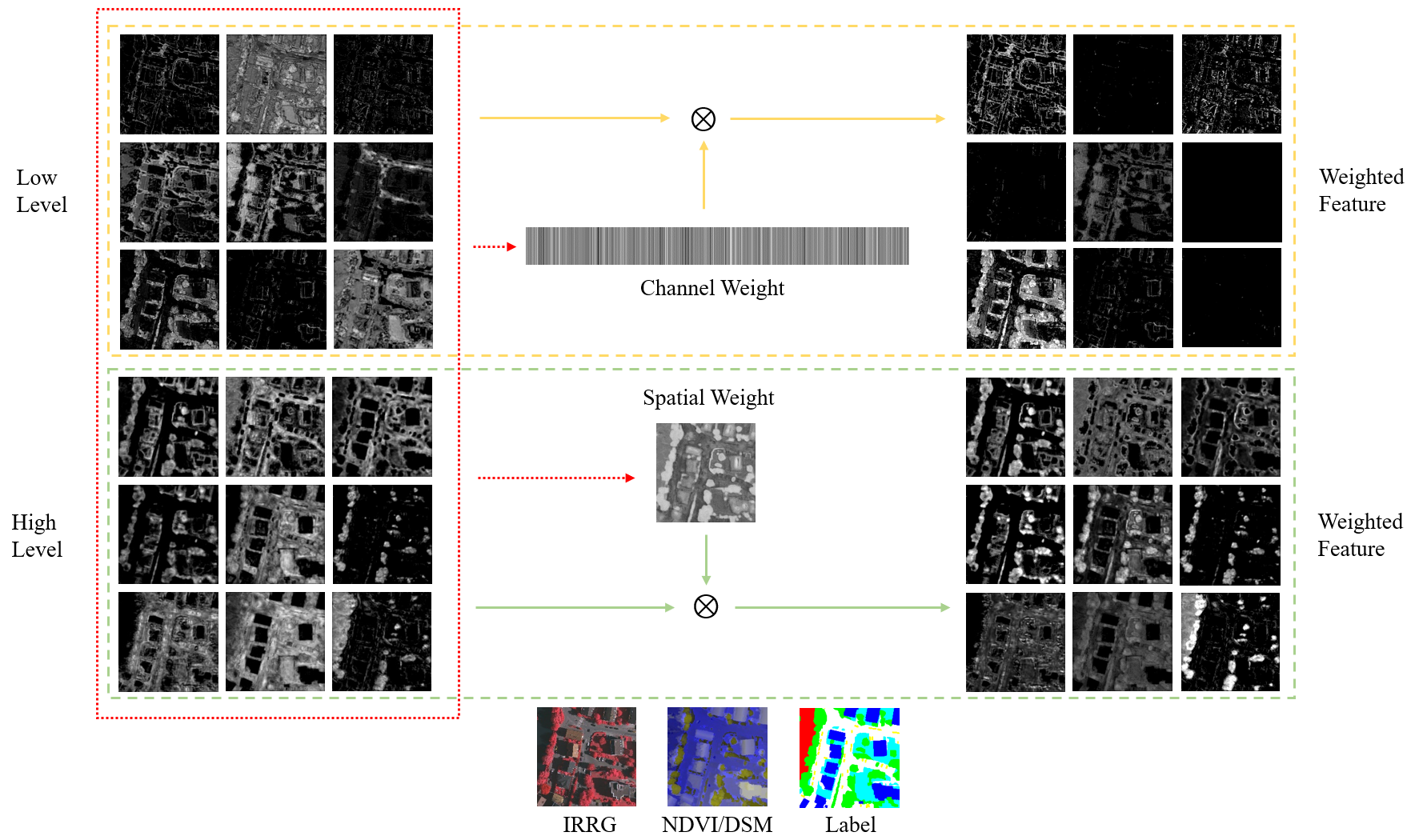}
\caption{Visualization of the feature maps in the RAFB module.}
\label{fig:rafbvis}
\end{figure}

Figure \ref{fig:rafbvis} shows that the RAFB module simultaneously learns the CA weight and the SA weight from the simply fused features. The CA weight has the abstract expression ability to filter the low-level feature maps to maintain helpful contextual information for classification. The SA weight has rich position information to focus the network on the brighter area in the high-level features and optimize the spatial features. The RAFB module uses the mutual restriction of the high-level abstract features and the low-level spatial features to refine the useful contextual information of the fused features. 

\subsection{Attention-Fused Network}

With the DFN as the baseline architecture, the MPE as the multipath inputs encoder, the MAFB as the feature fusion module for multipath features, and the RAFB as the feature fusion module for different level features, we propose the attention-fused network (AFNet) architecture, as shown in Figure \ref{fig:afnet:main}. 

To refine the feature, we use deep supervision to obtain a better performance and make AFNet easier to optimize. We choose cross-entropy loss to supervise each stage's outputs of the decoder. The loss value is used to quantify the difference between the forward propagation result of the network and the ground truth of the samples. The smaller the loss value is, the closer the forward propagation result is to the ground truth value, and the closer the parameters of the network are to convergence. The Adam optimizer takes the loss function as the optimization goal and places the loss value as close to 0 as possible. The cross-entropy loss is calculated by 

\begin{equation}
J=\frac{1}{N}\sum_{n=1}^{N}[y_n\log \hat{y}_n+(1-y_n)\log (1-\hat{y}_n)]
\label{eq:celoss}
\end{equation}

\noindent
where $N$ represents the total number of samples, $y_n$ represents the probability that the ground truth is true, $1-y_n$ represents the probability that the ground truth is false, $\hat{y}_n$ represents the probability that the forward propagation result is true, and $1-\hat{y}_n$ represents the probability that the forward propagation result is false. 

Additionally, there are other commonly used loss functions, such as focal loss \cite{lin2017focal} and dice loss. Focal loss is used to solve the hard sample problem, while dice loss directly uses intersection over union (IoU) as the optimization goal. On the ISPRS Vaihingen 2D dataset, the performance of the network is relatively stable, and the parameters of the network can converge normally. The accuracy of the car category is 0 if we use focal loss. Convergence does not occur if we use dice loss. A similar phenomenon appears on the ISPRS Potsdam 2D dataset. Therefore, we ultimately choose cross-entropy loss as the loss function to train our proposed AFNet. 

\begin{figure}[ht]
\centering
\includegraphics[width=0.5\linewidth]{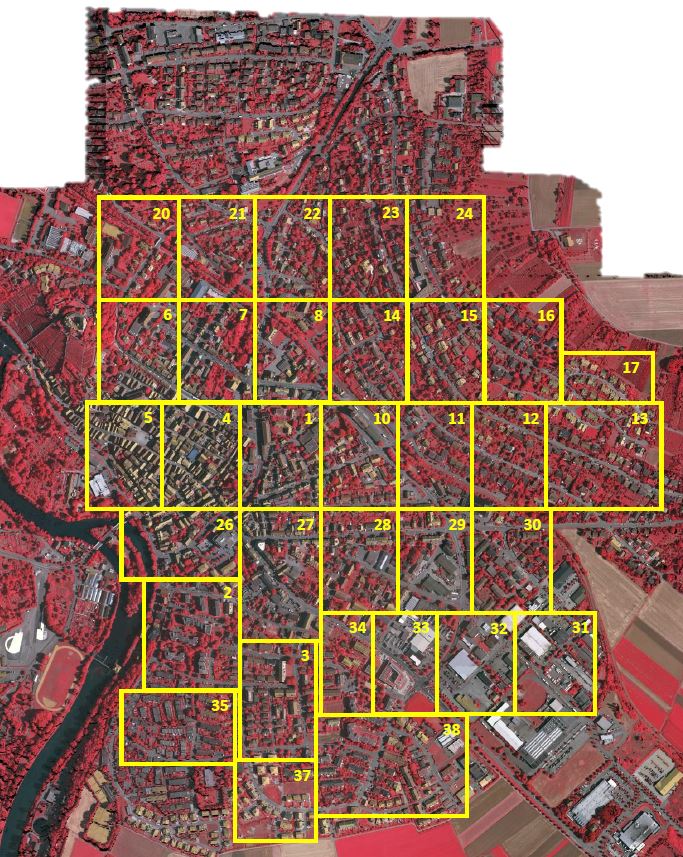}
\caption{Overview of the ISPRS Vaihingen 2D dataset. There are 33 tiles of true ortho photos. The number in the upper right corner of each tile represents the ID number of the tile. Figure source: \url{https://www2.isprs.org/commissions/comm2/wg4/benchmark/2d-sem-label-vaihingen/} (accessed on 29th March 2021).}
\label{fig:vaihingen_dataset}
\end{figure}

\section{Experimental Results}
\label{sec:4}

\subsection{Datasets}

\subsubsection{ISPRS Vaihingen 2D Dataset}

The ISPRS Vaihingen 2D dataset is a benchmark dataset of aerial remote sensing images labeled by the International Society for Photogrammetry and Remote Sensing (ISPRS). The dataset contains six types of land-cover categories, namely, impervious surfaces (imp\_surf), buildings, low vegetation (low\_veg), trees, cars, and clutter/background (clutter). The Vaihingen dataset contains aerial remote sensing images taken by drone in Vaihingen town, Germany. As shown in Figure \ref{fig:vaihingen_dataset}, there are 33 tiles of true ortho photo (TOP), which consists of the near-infrared (IR) channel, red (R) channel, and green (G) channel. Corresponding DSM data are also provided. The DSM represents the elevations of trees, buildings, and other target objects. We use IRRG and DSM data for training and inference. 

The average size of these tiles is $2494\times 2064$ pixels, and the spatial resolution is 9 cm. A total of 17 tiles are used for online evaluation. The dataset provider recently disclosed the labels of this part of the samples and provided a C++ program for accuracy evaluation. Therefore, this part of the samples is used as the test set to evaluate the accuracy of the networks. The earlier 16 samples are divided into the training set and validation set according to different proportions, which are adjusted at two different stages. In the stage of network design and debugging, two samples (IDs 1 and 13) are selected as the validation set, and the rest of the samples are used as the training set. After the network becomes stable, all 16 samples are used as the training set, and no independent validation set is used. 

\begin{figure}[ht]
\centering
\includegraphics[width=0.7\linewidth]{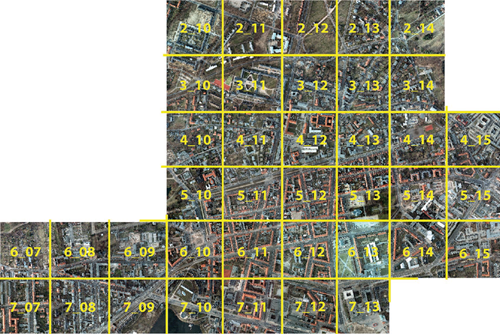}
\caption{Overview of the ISPRS Potsdam 2D dataset. There are 38 tiles of true ortho photos. The number in the center of each tile represents the ID number of the tile. Figure source: \url{https://www2.isprs.org/commissions/comm2/wg4/benchmark/2d-sem-label-potsdam/} (accessed on 29th March 2021).}
\label{fig:potsdam_dataset}
\end{figure}

\subsubsection{ISPRS Potsdam 2D Dataset}

The ISPRS Potsdam 2D dataset is a benchmark dataset of aerial remote sensing image labels provided by the ISPRS. The dataset contains six types of land-cover categories, namely, impervious surfaces (imp\_surf), buildings, low vegetation (low\_veg), trees, cars, and clutter/background (clutter). The Potsdam dataset contains aerial remote sensing images taken by drone in Potsdam city, Germany. As shown in Figure \ref{fig:potsdam_dataset}, there are 38 tiles of TOPs, which consist of the near-infrared (IR) channel, red (R) channel, green (G) channel, and blue (B) channel. Corresponding DSM data are also provided. The DSM represents the elevations of trees, buildings, and other target objects. We use IRRGB and DSM data for training and inference. 

The size of all these tiles is $6000\times 6000$ pixels, and the spatial resolution is 5 cm. A total of 14 tiles are used for online evaluation. The dataset provider recently disclosed the labels of this part of the samples and provided a C++ program for accuracy evaluation. Therefore, this part of the samples is used as the test set to evaluate the accuracy of the networks. The earlier 24 samples are divided into the training set and validation set according to different proportions, which are adjusted at two different stages. In the stage of network design and debugging, four samples (IDs 2\_11, 4\_11, 6\_9, and 6\_11) are selected as the validation set, and the rest of the samples are used as the training set. After the network becomes stable, all 24 samples are used as the training set, and no independent validation set is used. 

\begin{figure}[ht]
\centering
\includegraphics[width=0.5\linewidth]{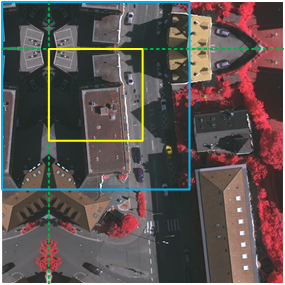}
\caption{Slicing the image with a 50\% overlap. The blue box indicates the slice range, the yellow box indicates the actual valid inference range, and the green dashed line indicates the mirror axis.}
\label{fig:overlap}
\end{figure}

\subsection{Implementation Details}

\subsubsection{Data Preprocessing}

The optical data of the Vaihingen dataset are IRRG data, and the optical data of the Potsdam dataset are IRRGB data. For convenience, we use the IRRG data to represent the optical data. However, we should remember that there is another blue channel in the Potsdam dataset.

According to Equation \ref{eq:norm}, we normalize the IRRG data and DSM data to speed up the model convergence. 

\begin{equation}
{image}_{\rm norm}=\frac{{image}_{\rm origin}-{mean}}{{std}}
\label{eq:norm}
\end{equation}

\noindent
where ${image}_{\rm norm}$ represents the normalized data, ${image}_{\rm origin}$ represents the original data, ${mean}$ represents the mean value of the corresponding channel in the original data, and ${std}$ represents the standard deviation of the corresponding channel in the original data. 

\begin{equation}
{ndvi}=\frac{{nir}-r}{{nir}+r}
\label{eq:ndvi}
\end{equation}

In addition to the IRRG and DSM data provided in the dataset, we also use NDVI data. According to Equation \ref{eq:ndvi}, NDVI data can be calculated from the IR and R channels in the IRRG data. NDVI data do not require normalization because they are distributed between $-1$ and $1$. After processing, we divide the data into two groups. One group is the IRRG image data, and the other group is the NDVI/DSM auxiliary data. 

Due to the GPU memory limitation, the size of each tile is too large to be directly fed into the GPU. Therefore, we first slice the IRRG image data and NDVI/DSM auxiliary data. Because the semantic information at the slice boundary is not complete, the prediction accuracy is low in this part, which results in obvious edge effects after stitching the slices back together. Therefore, we add a certain degree of overlap when slicing. Increasing the degree of overlap will greatly increase the sliced image data and increase the inference time overhead. To avoid edge effects as much as possible and to control the inference time within an acceptable range, we set the overlap to 50\%. This slicing method with a degree of overlap still cannot solve the accuracy degradation of the outermost part of the image. Therefore, we first perform the mirror expansion process on the outermost part of the original image. Then, we discard the less accurate mirrored part. As shown in Figure \ref{fig:overlap}, the blue box indicates the slice range, the yellow box indicates the actual valid inference range, and the green dashed line indicates the mirror axis. 

\subsubsection{Data Augmentation}

In the training stage, we use a data augmentation operation to increase the dataset, improve the generalization ability of the model, and avoid overfitting. The data augmentation methods we used are random horizontal flip, random vertical flip, random rotation, and random crop. 

In the inference stage, we also use a data augmentation operation, including horizontal flip, vertical flip, and rotation. This stage of data augmentation is also known as test-time augmentation (TTA). The TTA performs multiple inferences on the augmented data and integrates multiple inference results. 

\subsubsection{Training}

The proposed AFNet is implemented in the PyTorch deep learning framework \cite{paszke2019pytorch}. We use one NVIDIA TITAN Xp GPU for training, and the memory of the GPU is 12 GB. We find that for high-resolution images, the larger the slice is, the higher the accuracy is. To ensure that the GPU’s computing resources are fully utilized, we set the input size of the network to $640\times 640$ pixels. Since we use the random crop data augmentation operation, we set the slice size to $800\times 800$ pixels and the overlap to 400 pixels. The AFNet has two encoders, and the decoder is complicated, so the batch size is set to 2. We choose Adam as the optimizer with betas set to default values of 0.9 and 0.999, eps set to a default value $1\times 10^{-8}$, and weight decay set to $1\times 10^{-4}$. The learning rate uses the WarmUp strategy and Step strategy. The initial learning rate is set to $1\times 10^{-5}$. According to the WarmUp strategy (see Equation \ref{eq:warmup_lr}), the learning rate rises to $1\times 10^{-3}$ at the 100th epoch. Then, according to the Step strategy (see Equation \ref{eq:lr}), the learning rate multiplies by a factor of 0.1 every 200 epochs. The maximum iteration period is 1000 epochs. 

\begin{equation}
{lr}={lr}_0\cdot {(\frac{{lr}_1}{{lr}_0})}^{\frac{{current\_num\_iter}}{{total\_num\_iter}}}
\label{eq:warmup_lr}
\end{equation}

\noindent
where

\begin{equation}
{total\_num\_iter}={total\_num\_epoch}\times {num\_iter\_per\_epoch}
\label{eq:warmup_iter}
\end{equation}

\noindent 
where ${lr}$ represents the current learning rate, ${lr}_0$ represents the initial learning rate, ${lr}_1$ represents the learning rate at the end of the WarmUp strategy, ${current\_num\_iter}$ represents the current number of iterations, ${total\_num\_iter}$ represents the total number of iterations in the WarmUp strategy, ${total\_num\_epoch}$ represents the total number of epochs in the WarmUp strategy, and ${num\_iter\_per\_epoch}$ represents the number of iterations per epoch. 

\begin{equation}
{lr}'=\alpha \cdot {lr}
\label{eq:lr}
\end{equation}

\noindent
where ${lr}'$ represents the current learning rate, ${lr}$ represents the last learning rate, and $\alpha$ represents the factor in the Step strategy. 

\subsubsection{Inference}

In the inference stage, we set the input size of the network to $1920\times 1920$ pixels and the overlap to 960 pixels. We perform the TTA for all input images. Before the network performs the Argmax calculation, we add the inference result probabilistic feature maps output from multiple augmented data to obtain a new combined probabilistic feature map. Then, the Argmax calculation is performed to obtain the classification result. The TTA can significantly improve the accuracy of the inference results, but it will also exponentially increase the inference time. 

\subsection{Evaluation Metrics}

The overall accuracy (OA) is defined in Equation \ref{eq:oa}. The OA is the ratio of the number of correctly classified pixels to the total number of pixels. 

\begin{equation}
{OA}=\frac{{num\_pixels}_{\rm correct}}{{num\_pixels}_{\rm total}}
\label{eq:oa}
\end{equation}

\noindent
where ${num\_pixels}_{\rm correct}$ represents the number of correctly classified pixels and ${num\_pixels}_{\rm total}$ represents the total number of pixels. 

The accuracy of each category is evaluated using the F1 score. The F1 score (see Equation \ref{eq:f1}) is calculated by precision (see Equation \ref{eq:precision}) and recall (see Equation \ref{eq:recall}). In the confusion matrix, true positives (TPs) are the elements on the main diagonal, false positives (FPs) are the sum of the elements in each column except the elements on the main diagonal, and false negatives (FNs) are the sum of the elements in each row except the elements on the main diagonal. 

\begin{equation}
F_1=2\cdot \frac{{precision}\cdot {recall}}{{precision}+{recall}}
\label{eq:f1}
\end{equation}

\noindent
where

\begin{equation}
{precision}=\frac{{TP}}{{TP}+{FP}}
\label{eq:precision}
\end{equation}

\begin{equation}
{recall}=\frac{{TP}}{{TP}+{FN}}
\label{eq:recall}
\end{equation}

We notice that the OA is not adequately sensitive to the small categories. Therefore, in addition to the evaluation metrics recommended by the ISPRS, we use the mean F1 score for an overall evaluation. The mean F1 score is the average F1 score of each category. 

\subsection{Experiments on the Vaihingen Dataset}

\subsubsection{Ablation Study}

\begin{figure}[H]
\centering
\includegraphics[width=0.63\linewidth]{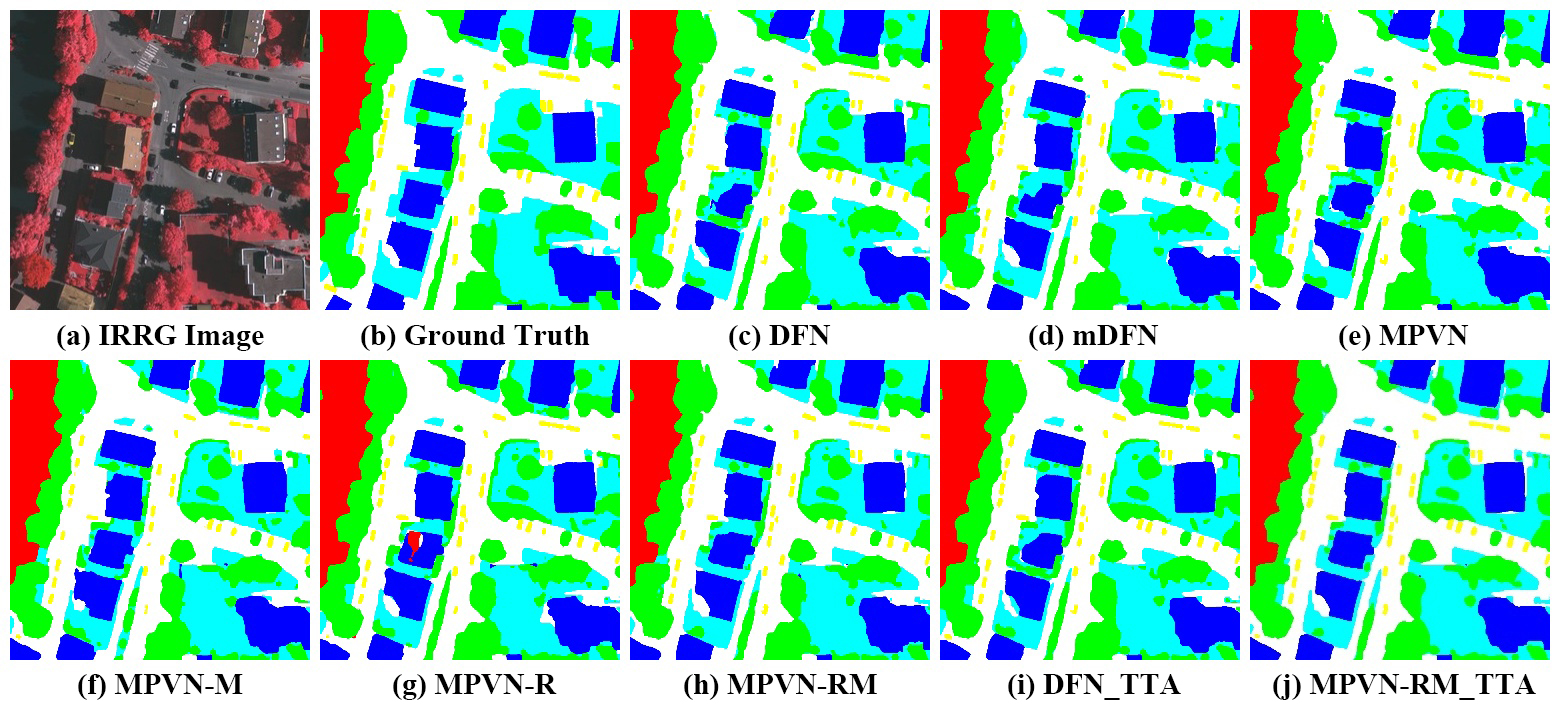}
\vfill
\includegraphics[width=0.63\linewidth]{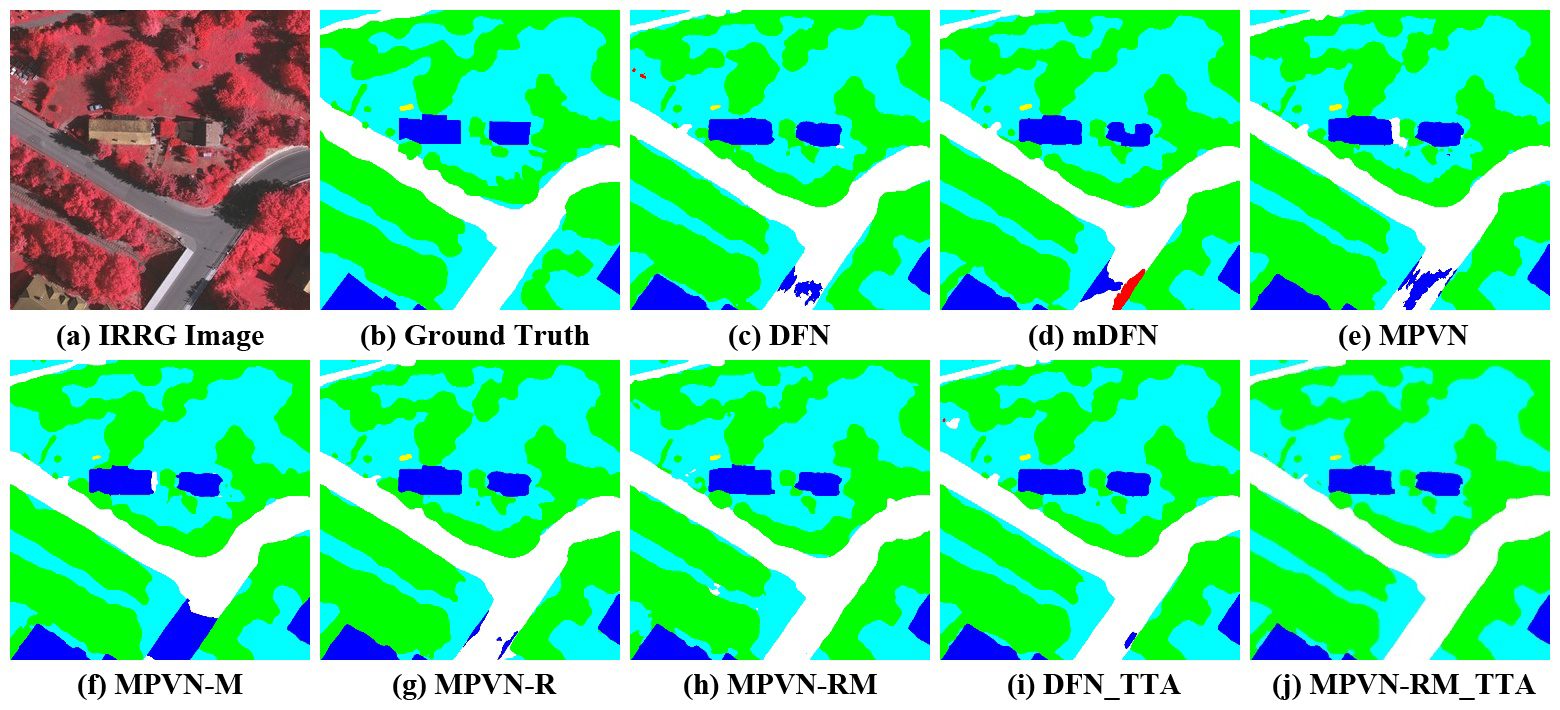}
\vfill
\includegraphics[width=0.63\linewidth]{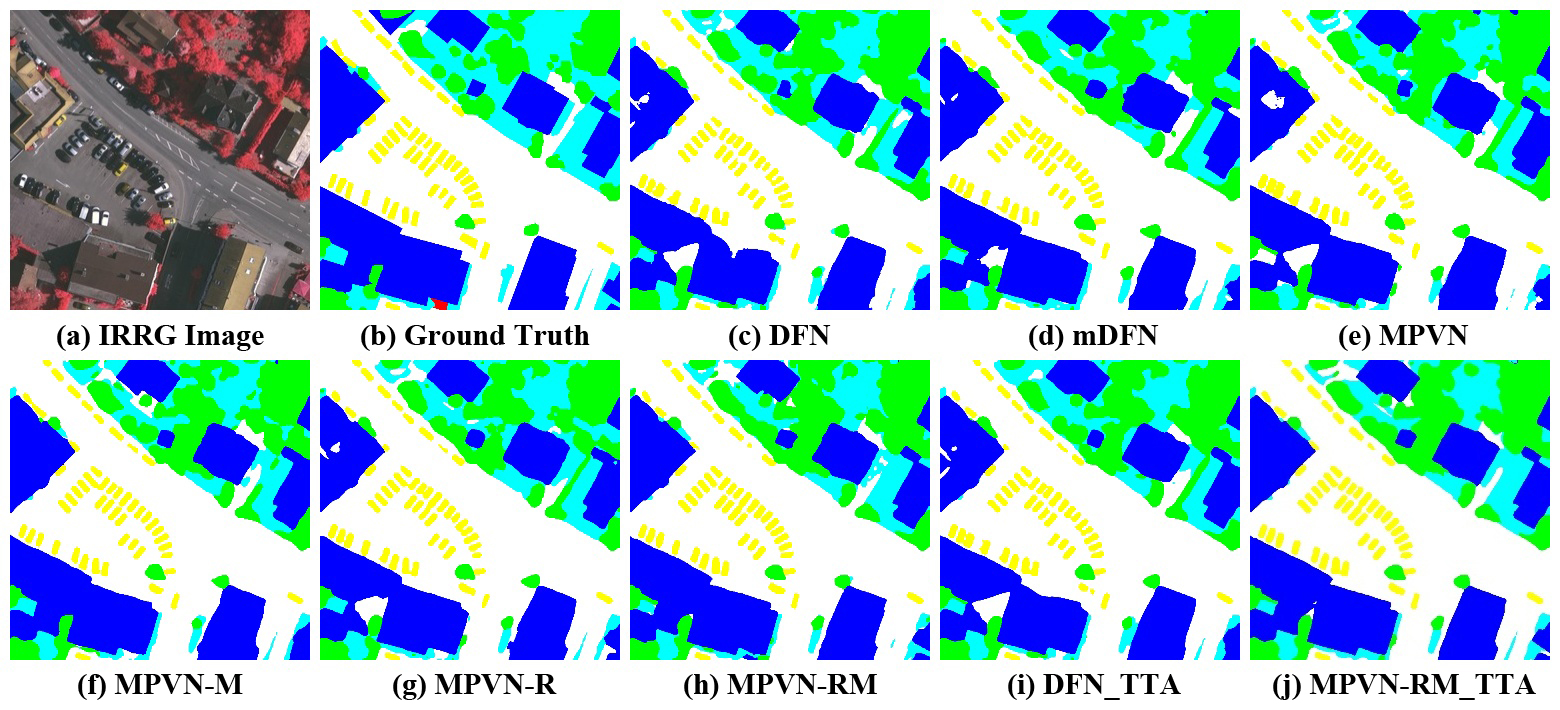}
\vfill
\includegraphics[width=0.63\linewidth]{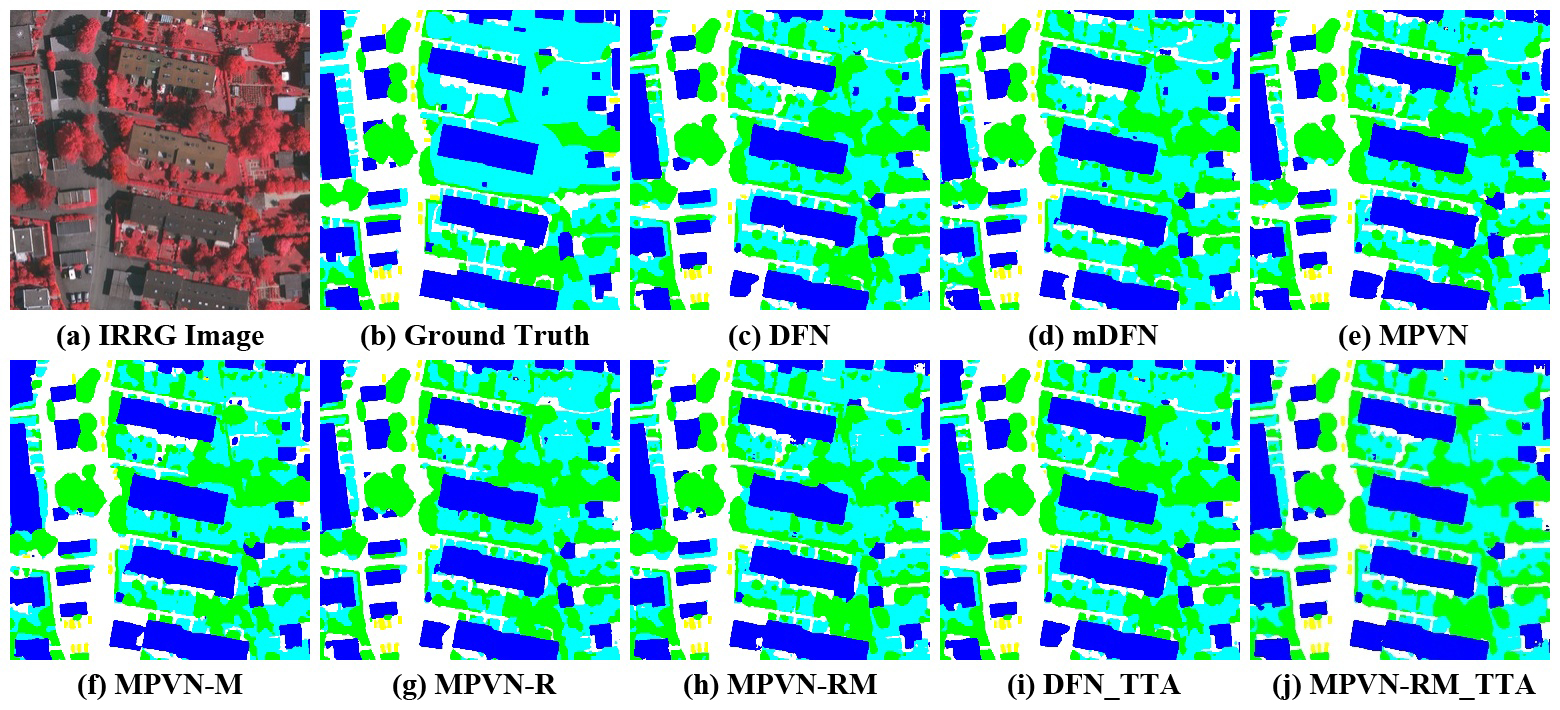}
\vfill
\includegraphics[width=0.32\linewidth]{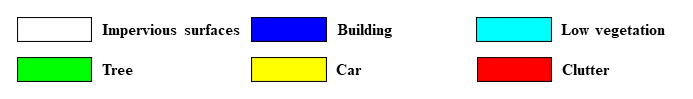}
\caption{Ablation study for our proposed AFNet on the ISPRS Vaihingen 2D dataset. (a) IRRG image. (b) Ground truth. Inference result of (c) the DFN, (d) the modified DFN with stacked data (mDFN), (e) the DFN with the MPE module (MPVN), (f) the MPVN with the MAFB module (MPVN-M), (g) the MPVN with the RAFB module (MPVN-R), (h) the MPVN with the RAFB module and the MAFB module (MPVN-RM), (i) the DFN with the TTA strategy (DFN\_TTA), (j) the MPVN-RM with the TTA strategy (MPVN-RM\_TTA).}
\label{fig:ablation_study}
\end{figure}

In this subsection, we gradually decompose the AFNet to show the effect of each module, which is proposed in this paper. Each experimental network architecture is evaluated with the ISPRS Vaihingen 2D dataset. We compare the accuracy of each architecture (see Table \ref{table:module}). Some examples of the results of the test set are shown in Figure \ref{fig:ablation_study}. Our proposed AFNet can effectively improve the classification accuracy. 

\begin{table}[ht]
\centering
\begin{tabular}{c c c c c c c c}
\hline
\textbf{Method} & \textbf{imp\_surf} & \textbf{building} & \textbf{low\_veg} & \textbf{tree} & \textbf{car}  & \textbf{OA}   & \textbf{Mean F1} \\
\hline
DFN             & 92.6               & 95.3              & 83.6              & 89.3          & 87.2          & 90.4          & 89.60            \\
mDFN            & 92.2               & 95.6              & 83.9              & 89.5          & 87.0          & 90.5          & 89.64            \\
MPVN            & \textbf{92.8}      & 95.6              & 83.7              & 89.3          & 86.7          & 90.6          & 89.62            \\
MPVN-M          & 92.4               & \textbf{96.1}     & 84.3              & 89.5          & 87.0          & 90.8          & 89.86            \\
MPVN-R          & \textbf{92.8}      & 96.0              & 84.6              & \textbf{90.0} & 87.1          & 91.0          & 90.10            \\
MPVN-RM         & 92.7               & \textbf{96.1}     & \textbf{84.8}     & 89.9          & \textbf{87.9} & \textbf{91.1} & \textbf{90.28}   \\
\hline
\end{tabular}
\caption{The effect of the MPE module, the RAFB module and the MAFB module on the ISPRS Vaihingen 2D dataset.}
\label{table:module}
\end{table}

\textbf{Baseline.} We choose the DFN without the border network as the baseline network and choose ResNet-50 as the encoder. The input data are the IRRG data. The OA of the baseline network for the test set is 90.4\%, and the mean F1 score is 89.6\%. To compare the effect of the MPE module, we simply stack the NDVI/DSM auxiliary data to the IRRG data and train a modified DFN model. This modified DFN changes the number of input channels of the first convolutional layer to the number of channels of the stacked data. We name this network architecture mDFN. The OA of the mDFN for the test set is 90.5\%, and the mean F1 score is 89.64\%.

\textbf{Multipath Encoder.} We replace the encoder in the DFN with the MPE module, where the main branch is ResNet-50, and the auxiliary branch is ResNet-18. We feed the IRRG image data and NDVI/DSM auxiliary data into the main branch and the auxiliary branch, respectively. The features extracted from the two branches of the encoder are directly added to obtain the fused features. Then, the fused feature is fed into the decoder of the DFN. We name this network architecture MPVN. The OA of the MPVN for the test set is 90.6\%, and the mean F1 score is 89.62\%. 

\textbf{Multipath Attention-Fused Block.} We replace the addition operation for fusing two branches of features in the MPVN encoder with the MAFB module proposed in this paper. The image features and auxiliary features are fused by the MAFB module and fed into the MPVN decoder. We name this network architecture the MPVN-M. The OA of the MPVN-M for the test set is 90.8\%, and the mean F1 score is 89.86\%. 

\textbf{Refinement Attention-Fused Block.} We replace the CAB module in the DFN decoder with the RAFB module proposed in this paper and use the RAFB to fuse the high-level abstract features and the low-level spatial features. We name this network architecture MPVN-R. The OA of the MPVN-R for the test set is 91.0\%, and the mean F1 score is 90.1\%. 

\textbf{Attention-Fused Network.} We simultaneously apply the MAFB module and RAFB module to the MPVN architecture. The image features and the auxiliary features are fused by the MAFB module. The high-level abstract features and the low-level spatial features are fused by the RAFB module. We name this network architecture MPVN-RM. The OA of the MPVN-RM for the test set is 91.1\%, and the mean F1 score is 90.28\%. The MPVN-RM is actually our proposed AFNet. 

\textbf{Test-Time Augmentation.} The MPVN-RM with the MPE module, RAFB module, and MAFB module can significantly improve the inference accuracy. We apply the TTA strategy to each network mentioned above in the inference stage, and the performance of the network is further improved (see Table \ref{table:tta}). The OA of each network increases by approximately 0.4\% to 0.6\%. In particular, the OA of our proposed AFNet increases by 0.6\% to 91.7\%. 

\begin{table}[ht]
\centering
\begin{tabular}{c c c c c c c c}
\hline
\textbf{Method} & \textbf{imp\_surf} & \textbf{building} & \textbf{low\_veg} & \textbf{tree} & \textbf{car}  & \textbf{OA}   & \textbf{Mean F1} \\
\hline
DFN+TTA         & 93.0               & 95.6              & 84.4              & 89.8          & 88.1          & 90.9          & 90.18            \\
mDFN+TTA        & 92.7               & 95.8              & 84.7              & 89.9          & 88.6          & 91.0          & 90.34            \\
MPVN+TTA        & \textbf{93.2}      & 95.8              & 84.8              & 90.0          & 87.5          & 91.1          & 90.26            \\
MPVN-M+TTA      & 92.7               & 96.3              & 85.0              & 90.0          & 87.6          & 91.2          & 90.32            \\
MPVN-R+TTA      & 93.1               & 96.2              & 85.2              & 90.3          & 87.7          & 91.4          & 90.50            \\
MPVN-RM+TTA     & 93.1               & \textbf{96.5}     & \textbf{85.8}     & \textbf{90.6} & \textbf{88.8} & \textbf{91.7} & \textbf{90.96}   \\
\hline
\end{tabular}
\caption{The effect of the TTA strategy on the ISPRS Vaihingen 2D dataset.}
\label{table:tta}
\end{table}

\begin{figure}[ht]
\centering
\includegraphics[width=0.8\linewidth]{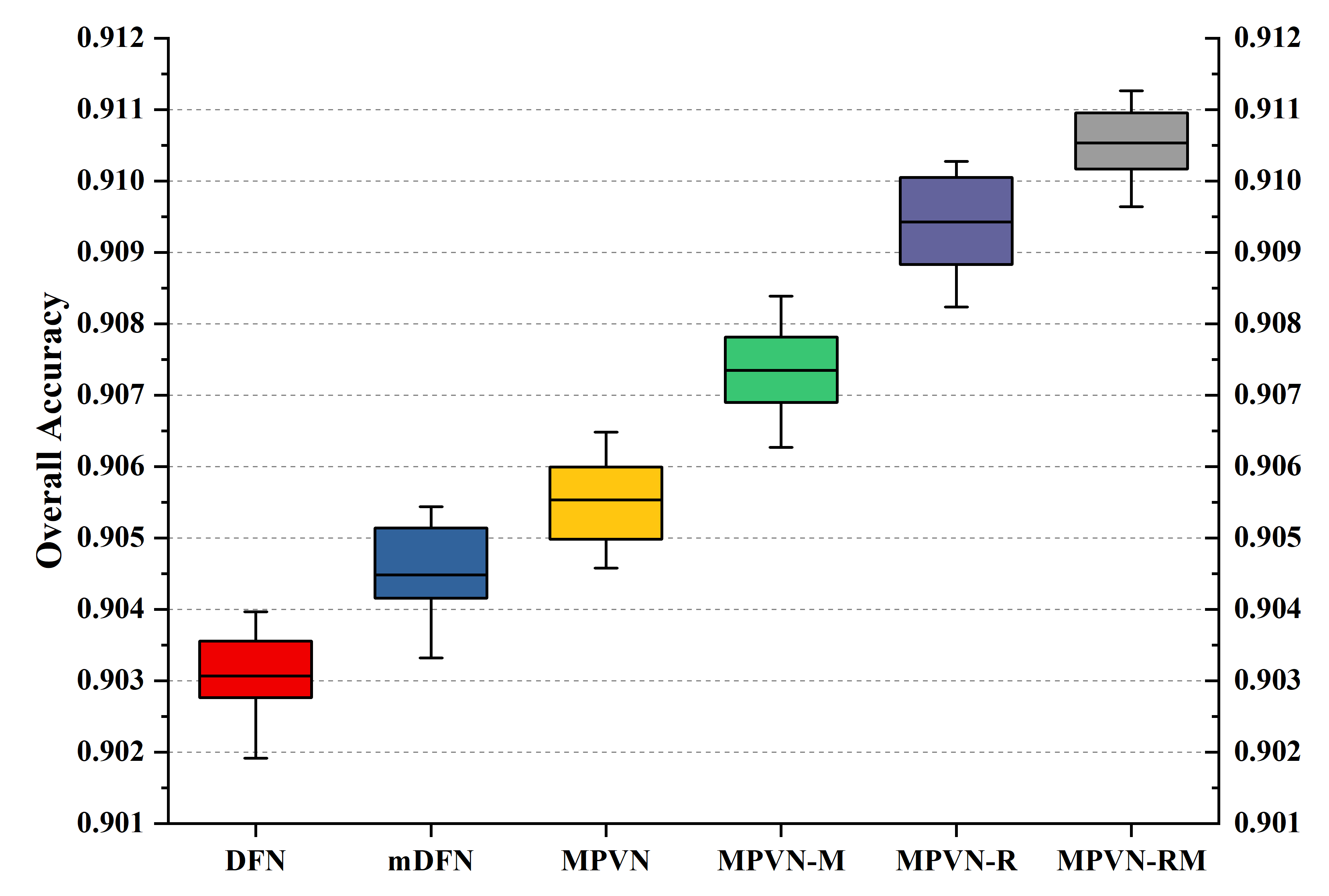}
\caption{Accuracy range, mean, and interquartile range (IQR) of each method for multiple runs.}
\label{fig:acc_mean_box}
\end{figure}

Since random errors occur during each training process, the inference accuracy fluctuates, even if the training parameters are exactly the same. We ran ten training sessions for each method and performed inference and accuracy evaluations on the test set without using the TTA strategy. As shown in Table \ref{table:mean_stddev}, we calculated the mean and standard deviation of the inference accuracy of each method for multiple runs. The accuracy range, mean, and interquartile range (IQR) of multiple runs are shown in Figure \ref{fig:acc_mean_box}. 

Using the C++ program provided by the organizer, we evaluated the accuracy of the AFNet’s inference results and generated a detailed evaluation webpage for the ISPRS Vaihingen 2D dataset. The evaluation webpage includes the following information: OA, individual accuracy of each category, individual accuracy of each image tile, confusion matrix, and the red-green image for showing the areas of a wrong classification. We uploaded this evaluation webpage to our web server. The webpage is available online at \url{http://research.yangxuan.me/isprs/vaihingen/radi/index.html} (accessed on 29th March 2021). 

\begin{table}[ht]
\centering
\begin{tabular}{c c c c c c c}
\hline
                & \textbf{DFN} & \textbf{mDFN} & \textbf{MPVN} & \textbf{MPVN-M} & \textbf{MPVN-R}  & \textbf{MPVN-RM} \\
\hline
\textbf{mean}   & 90.31        & 90.45         & 90.55         & 90.73           & 90.94            & 91.05            \\
\textbf{stddev} & 0.0586       & 0.0665        & 0.0611        & 0.0625          & 0.0673           & 0.0487           \\
\hline
\end{tabular}
\caption{Mean and standard deviation of the inference accuracy of each method for multiple runs.}
\label{table:mean_stddev}
\end{table}

In Figure \ref{fig:ablation_study}, we clearly see the effect of the MPE module, the RAFB module, the MAFB module and the TTA strategy. From the first group of results, we find that there is a building in the middle area of the picture, and there are apparent differences in the results under different methods. In the DFN, the result of the building misses a corner. In the mDFN, although we stacked the NDVI/DSM data with IRRG data for training, the missing corner remains there regardless of whether there is a slight improvement. In the MPVN, after adding the DSM data, the missing corner is significantly recovered. After adding the MAFB module, the missing corner almost disappeared because the features extracted from the NDVI/DSM data are effectively fused into the network. An incorrect classification area appeared only after we added the RAFB module to the MPVN because there are some conflicting multipath features without the MAFB module, which can suppress these features. With the addition of the MAFB module in the MPVN-R, the problem of misclassification is resolved. After using the TTA strategy, multiple inference results are integrated to eliminate random errors and improve the accuracy of the building boundary. 

According to the second group of results, a similar phenomenon exists in the road area. There are obvious errors in the roads extracted by the DFN and the mDFN, and the same problem still exists in the MPVN. However, we notice that the incorrectly classified flaws in the upper left corner of the picture have disappeared. In this case, the MAFB module introduces a new error in the road area, but its boundaries are very regular because the high-level abstract features are not effectively fused with the low-level spatial features without the RAFB module. After adding the RAFB module to eliminate inconsistent features, the incorrectly classified area on the road becomes significantly smaller. The road is flat, so the DSM data is important for road extraction. Therefore, after the MAFB module is used for the MPVN-R, the road area classification result is totally correct. Then, the TTA strategy refines the results and improves the OA. 

From the third group of results, the buildings in the bottom left corner, the upper left corner, and the bottom right corner of the picture show the improvement obtained by using our proposed AFNet architecture. These buildings have similar results in the DFN and the MPVN, with obvious errors. In the mDFN, the results are slightly better, but they are far from satisfactory. With the addition of the MAFB module in the MPVN, thanks to the DSM data, most of the errors disappeared. However, there is still an incorrectly classified area in the bottom right corner of the picture. With the addition of the RAFB module in the MPVN, the error result of the building in the upper left corner is basically resolved. After using the MAFB module in the MPVN-R, those three misclassification problems are completely resolved. Finally, we use the TTA strategy to repair some further details and improve the OA. 

In the fourth group of results, we find a similar problem to that in the previous examples in the building. After gradually adding the MPE structure, the RAFB module, and the MAFB module to the baseline DFN, the accuracy of the segmentation continues to improve. After using the TTA strategy in the end, many details have been fixed, and the accuracy has been significantly improved. With or without the TTA strategy, our proposed AFNet performs better than the baseline DFN. In summary, the MPE structure, the RAFB module, and the MAFB module proposed in this paper can progressively improve the semantic segmentation performance for very-high-resolution remote sensing imagery. 

\begin{table}[ht]
\centering
\begin{tabular}{c c c c c c c c}
\hline
\textbf{Method} 
& \textbf{imp\_surf} & \textbf{building} & \textbf{low\_veg} & \textbf{tree} & \textbf{car}  & \textbf{OA}   & \textbf{Mean F1} \\
\hline
UT\_Mev \cite{speldekamp2015automatic} 
& 84.3               & 88.7              & 74.5              & 82.0          & 9.9           & 81.8          & 67.88            \\
SVL\_3 \cite{gerke2014use} 
& 86.6               & 91.0              & 77.0              & 85.0          & 55.6          & 84.8          & 79.04            \\
DST\_2 \cite{sherrah2016fully} 
& 90.5               & 93.7              & 83.4              & 89.2          & 72.6          & 89.1          & 85.88            \\
UFMG\_4 \cite{nogueira2019dynamic} 
& 91.1               & 94.5              & 82.9              & 88.8          & 81.3          & 89.4          & 87.72            \\
ONE\_7 \cite{audebert2016semantic} 
& 91.0               & 94.5              & 84.4              & 89.9          & 77.8          & 89.8          & 87.52            \\
DLR\_9 \cite{marmanis2018classification} 
& 92.4               & 95.2              & 83.9              & 89.9          & 81.2          & 90.3          & 88.52            \\
DFN \cite{yu2018learning} 
& 92.6               & 95.3              & 83.6              & 89.3          & 87.2          & 90.4          & 89.60            \\
DFN+TTA 
& 93.0               & 95.6              & 84.4              & 89.8          & 88.1          & 90.9          & 90.18            \\
BKHN10 
& 92.9               & 96.0              & 84.6              & 89.8          & 88.8          & 91.0          & 90.42            \\
AFNet (t/v) 
& 92.8               & 96.4              & 84.6              & 89.2          & 89.1          & 91.0          & 90.42            \\
CASIA2 \cite{liu2018semantic} 
& \textbf{93.2}      & 96.0              & 84.7              & 89.9          & 86.7          & 91.1          & 90.10            \\
AFNet (t/n) 
& 92.7               & 96.1              & 84.8              & 89.9          & 87.9          & 91.1          & 90.28            \\
NLPR3 
& 93.0               & 95.6              & 85.6              & 90.3          & 84.5          & 91.2          & 89.80            \\
AFNet+TTA (t/v) 
& \textbf{93.2}      & \textbf{96.6}     & 85.5              & 89.8          & \textbf{89.3} & 91.5          & 90.88            \\
AFNet+TTA (t/n) 
& 93.1               & 96.5              & \textbf{85.8}     & \textbf{90.6} & 88.8          & \textbf{91.7} & \textbf{90.96}   \\
\hline
\end{tabular}
\caption{Accuracy comparisons between our AFNet and other state-of-the-art methods on the ISPRS Vaihingen 2D dataset.}
\label{table:methods}
\end{table}

\subsubsection{Comparing Methods}

\begin{figure}[H]
\centering
\includegraphics[width=0.71\linewidth]{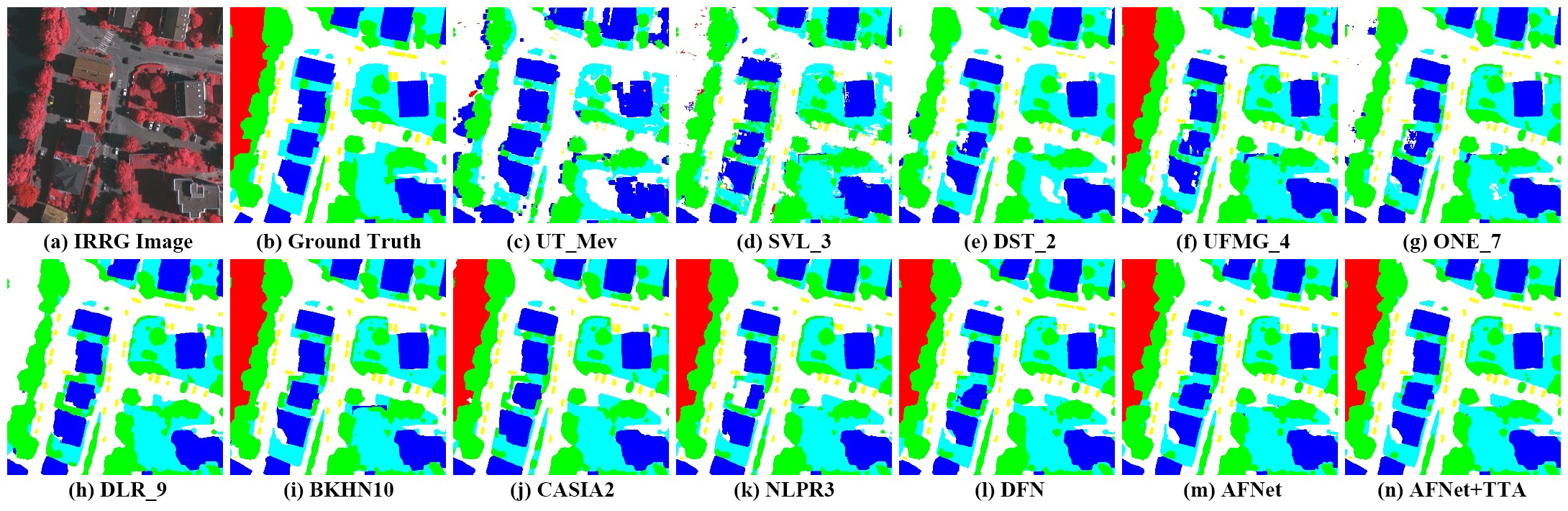}
\vfill
\includegraphics[width=0.71\linewidth]{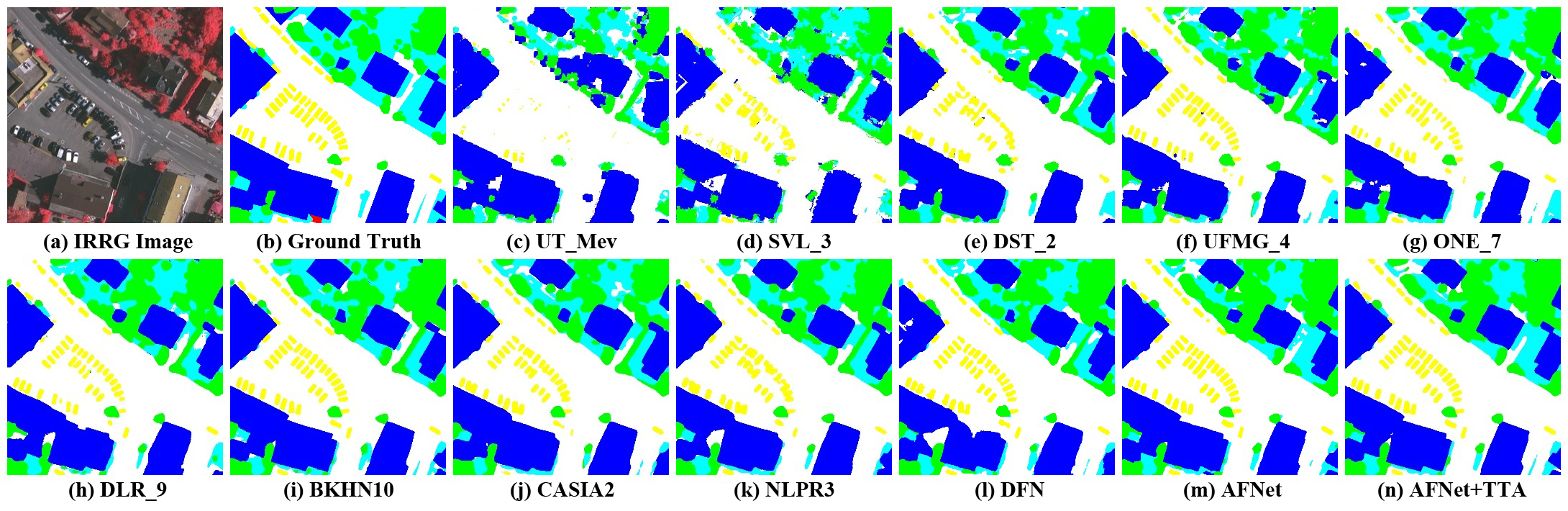}
\vfill
\includegraphics[width=0.71\linewidth]{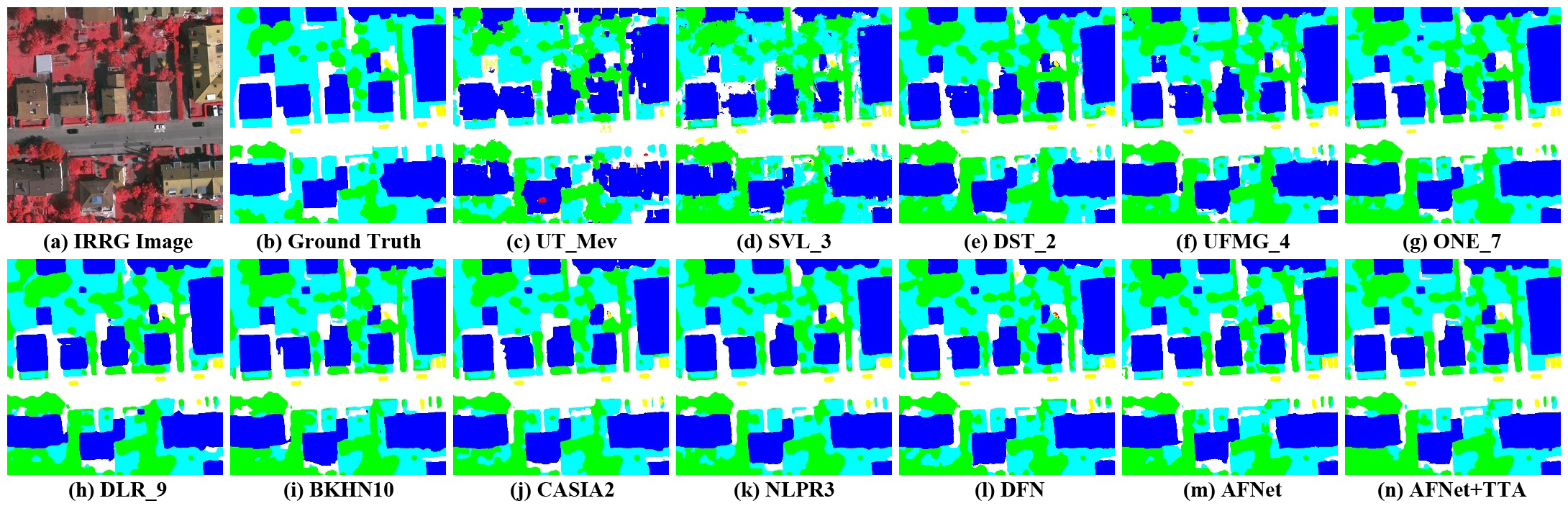}
\vfill
\includegraphics[width=0.71\linewidth]{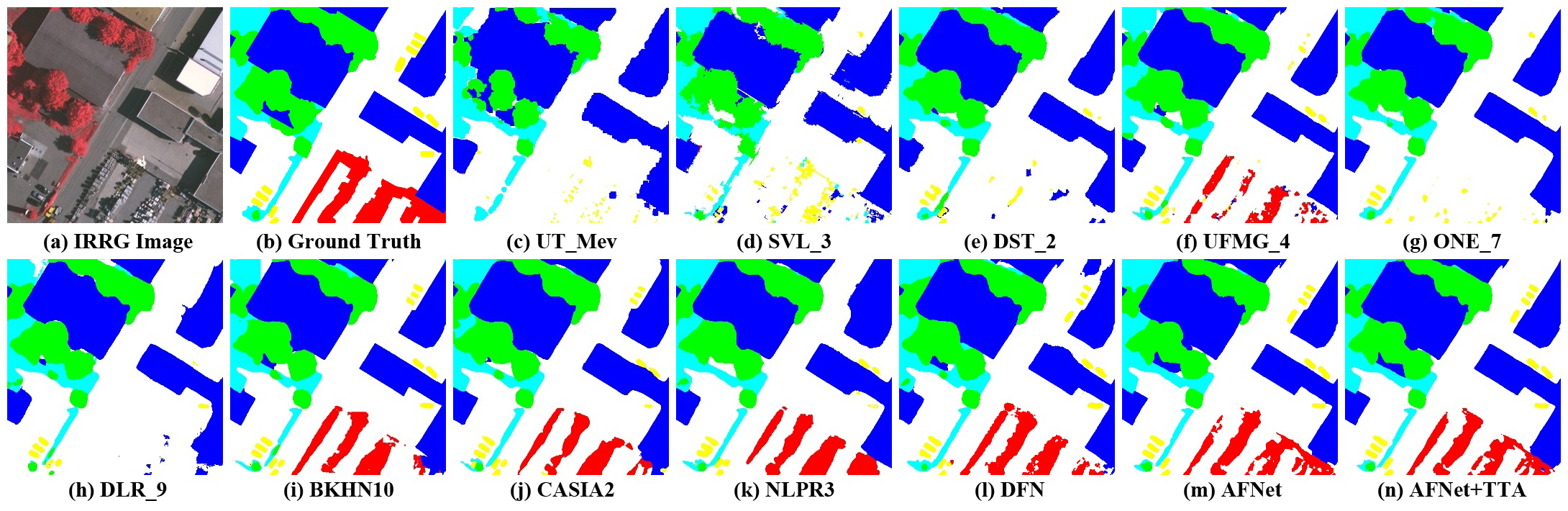}
\vfill
\includegraphics[width=0.71\linewidth]{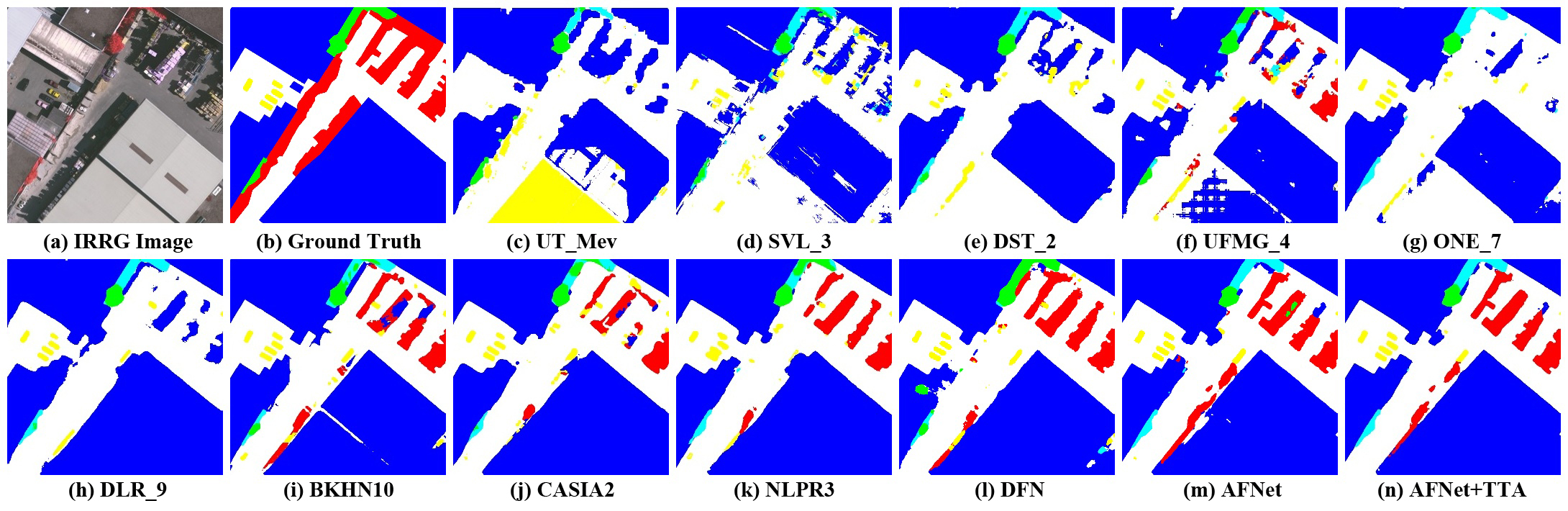}
\vfill
\includegraphics[width=0.31\linewidth]{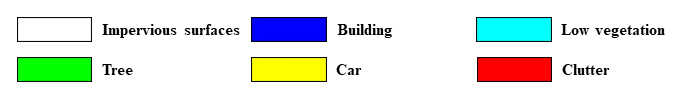}
\caption{Some examples of the results of the test set on the ISPRS Vaihingen 2D dataset. Comparisons between our AFNet and other state-of-the-art methods. (a) IRRG image. (b) Ground truth. Inference result of (c) UT\_Mev, (d) SVL\_3, (e) DST\_2, (f) UFMG\_4, (g) ONE\_7, (h) DLR\_9, (i) BKHN10, (j) CASIA2, (k) NLPR3, (l) the DFN, (m) our proposed AFNet, (n) our proposed AFNet with the TTA strategy.}
\label{fig:comparing_methods}
\end{figure}

Many state-of-the-art methods and results have been submitted to the ISPRS website \cite{webisprsresults}. The accuracy comparisons between our AFNet and those state-of-the-art methods are shown in Table \ref{table:methods}. Some examples of the test set results are shown in Figure \ref{fig:comparing_methods}. We see that our proposed AFNet achieves the best performance on the ISPRS Vaihingen 2D dataset. 

(1) AFNet: Our AFNet is a MPVN with the MPE module, the MAFB module, and the RAFB module. We use both IRRG data and DSM data for training and inference. NDVI data are also used as input data. We use the TTA strategy in the inference stage. Neither post-processing nor multimodel ensemble learning is used in AFNet. 

(2) UT\_Mev: Speldekamp et al. \cite{speldekamp2015automatic} proposed this method. This method is an unsupervised classification method. By calculating NDVI and combining DSM data, according to the characteristics of different categories of target objects, different NDVI thresholds and DSM thresholds are set for classification. The results are classified individually according to the categories. 

(3) SVL\_3: Gerke et al. \cite{gerke2014use} proposed this method. This method is based on SVL features, combined with the features of the NDVI, saturation, and DSM. The classifier uses a method based on AdaBoost and introduces the CRF algorithm to post-process the inference results. 

(4) DST\_2: Sherrah et al. \cite{sherrah2016fully} proposed this method. This network is based on the FCN, and it uses a hybrid structure to fuse DSM data and image data. The downsampling layer of the network is removed to retain the spatial position information. Finally, the CRF is introduced as the post-processing algorithm to refine the inference results. 

(5) UFMG\_4: Nogueira et al. \cite{nogueira2019dynamic} proposed this method. This network is based on a cascaded CNN, and it replaces the ordinary convolution operation with the dilated convolution operation. 

(6) ONE\_7: Audebert et al. \cite{audebert2016semantic} proposed this method. This network is based on the SegNet. Two encoders are used to extract IRRG features and DSM features. These two branch features are fused in the later stage of the decoder. 

(7) DLR\_9: Marmanis et al. \cite{marmanis2018classification} proposed this method. Multiple networks are used for ensemble learning, including the SegNet, VGG, and FCN. Both IRRG data and DSM data are used. An edge detection module is designed to improve the accuracy of training and inference. 

(8) BKHN10: This method is not published, and we only have a brief abstract. This network is based on FCN-8s and replaces the original encoder with ResNet-101. Multiple models are used for ensemble learning. Both IRRG data and DSM data are used for training and inference. 

(9) CASIA2: Liu et al. \cite{liu2018semantic} proposed this method. This network is based on the UNet. A self-cascaded CNN module is designed to fuse multiscale features. VGGNet and ResNet are the encoders of the network. Only IRRG data are used for training and inference. 

(10) NLPR3: This method is not published, and we only have a brief abstract. This network is based on the FCN. Fully connected conditional random fields (F-CRFs) are used to post-process the inference results. 

In Table \ref{table:methods}, we find that the overall performance of AFNet proposed in this paper is good. We use t/v to indicate that there is an independent validation set in the training stage. We use t/n to indicate that there is no independent validation set in the training stage. Not only is the OA the highest but also the accuracy of all categories of target objects is the highest in AFNet+TTA (t/v) and AFNet+TTA (t/n). BKHN10 uses the independent validation set in the training stage, but it uses five models for ensemble learning, which can improve accuracy. CASIA2 uses all data for training and no independent validation set. NLPR3 uses F-CRFs as the post-processing algorithm to refine the results. Our AFNet does not use a multimodel for ensemble learning and does not use any post-processing algorithm. We obtained 91.0\% OA for training with the independent validation set and 91.1\% OA for training without the independent validation set. These scores are almost the same as those of the nearest competitors, BKHN10, CASIA2, and NLPR3. With the TTA strategy, the accuracy of our proposed AFNet has significantly improved. Although the nearest competitors do not mention whether TTA is used or not, since TTA is a well-known and widely used technique, it is essential to use the TTA strategy to improve the final results. The proportion of impervious surfaces, buildings, low vegetation, and trees is relatively large, and the proportion of cars is very small. The OA is not affected when the accuracy of the car category is not high. However, the mean F1 score is more sensitive. The mean F1 score of AFNet is also the best score, indicating that all categories of target objects perform well, including the car category. 

According to Figure \ref{fig:comparing_methods}, we can clearly see that our proposed AFNet outperforms all other state-of-the-art methods. There are many misclassifications in the UT\_Mev result because it is based on an unsupervised classification method. Such methods are generally inferior to supervised classification. There are many fragmentation errors in the SVL\_3 result because this method is a machine learning method, and its generalization performance is not as good as that of the deep learning method. The DST\_2 method uses eight times upsample, resulting in lower accuracy in the car and clutter categories. The UFMG\_4 method, ONE\_7 method, and DLR\_9 method do not use the global context and attention structure. These results have certain misclassification problems, and the clutter cannot be classified at all. Most state-of-the-art methods use both IRRG data and DSM data, except CASIA2 method. The accuracy of the segmentation results predicted by the BKHN10 method and CASIA2 method is high. However, there are still some flaws in the details because the attention structure is not used in these two methods. The NLPR3 method applies F-CRFs as the post-processing for segmentation results. However, the F-CRF has certain side effects on small target objects, resulting in a decrease in the accuracy of the car category. Our proposed AFNet uses the MPE structure, the MAFB module, and the RAFB module, is trained and inferred with both IRRG data and DSM data, and solves various problems mentioned above. Our AFNet achieves state-of-the-art performance on the ISPRS Vaihingen 2D dataset, demonstrating the superiority of our designed network structure. 

\subsection{Experiments on the Potsdam Dataset}

\begin{figure}[H]
\centering
\includegraphics[width=0.63\linewidth]{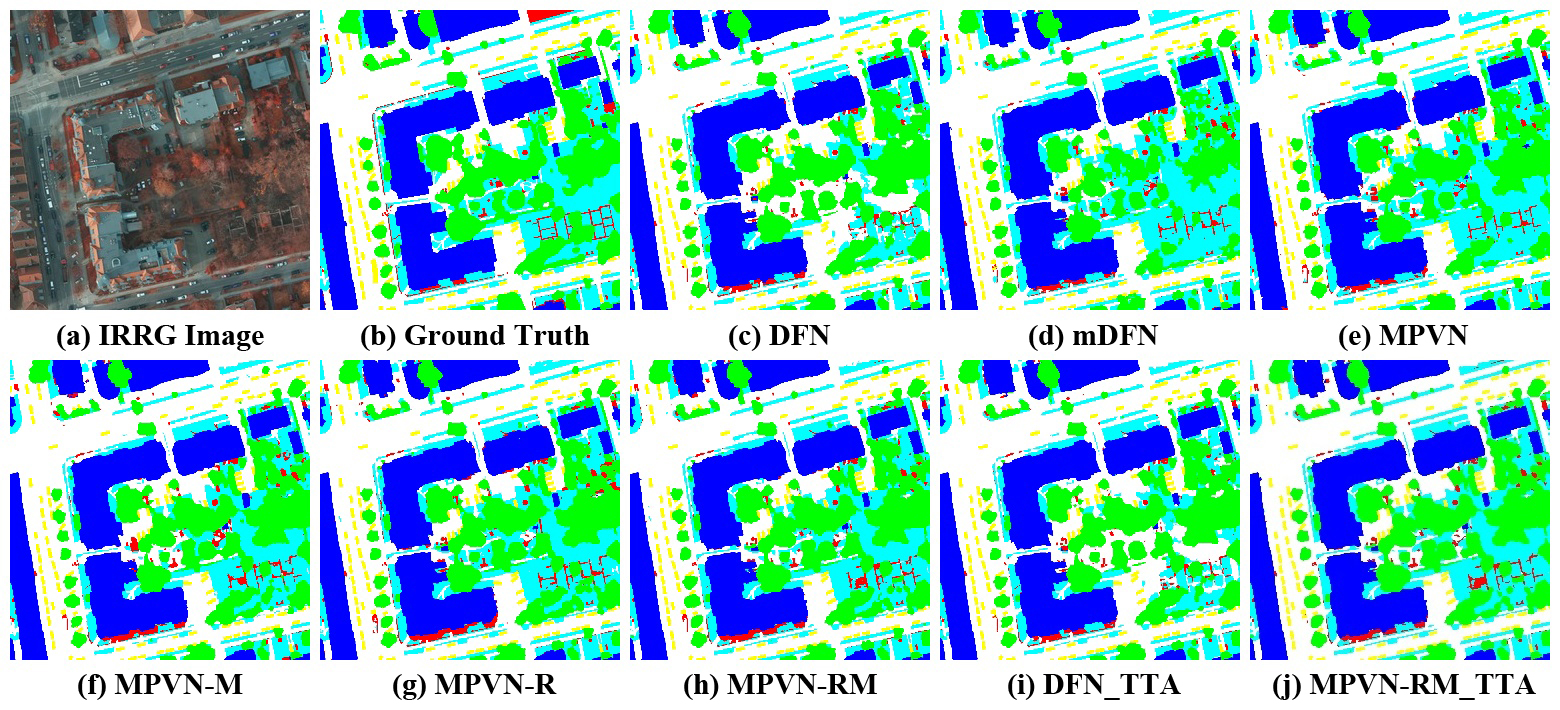}
\vfill
\includegraphics[width=0.63\linewidth]{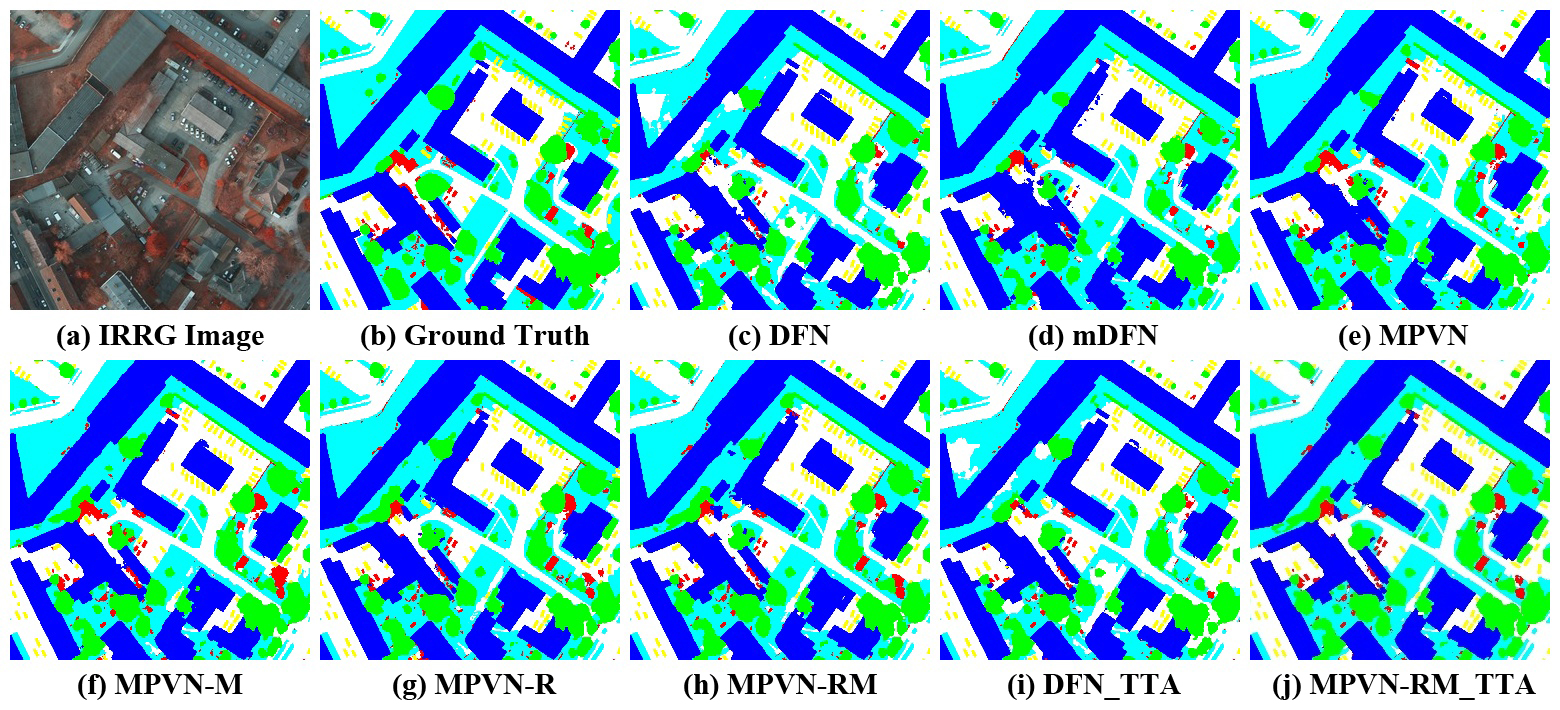}
\vfill
\includegraphics[width=0.63\linewidth]{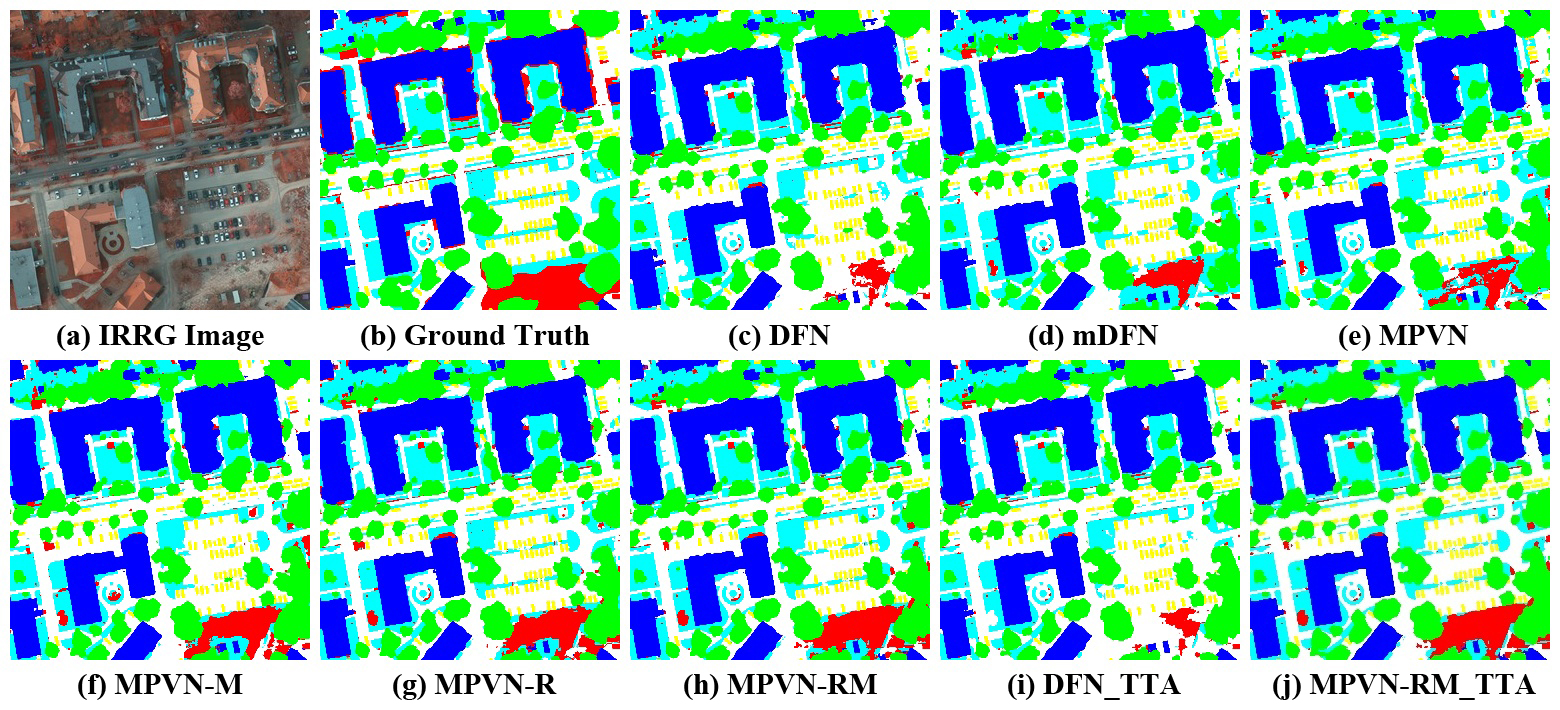}
\vfill
\includegraphics[width=0.63\linewidth]{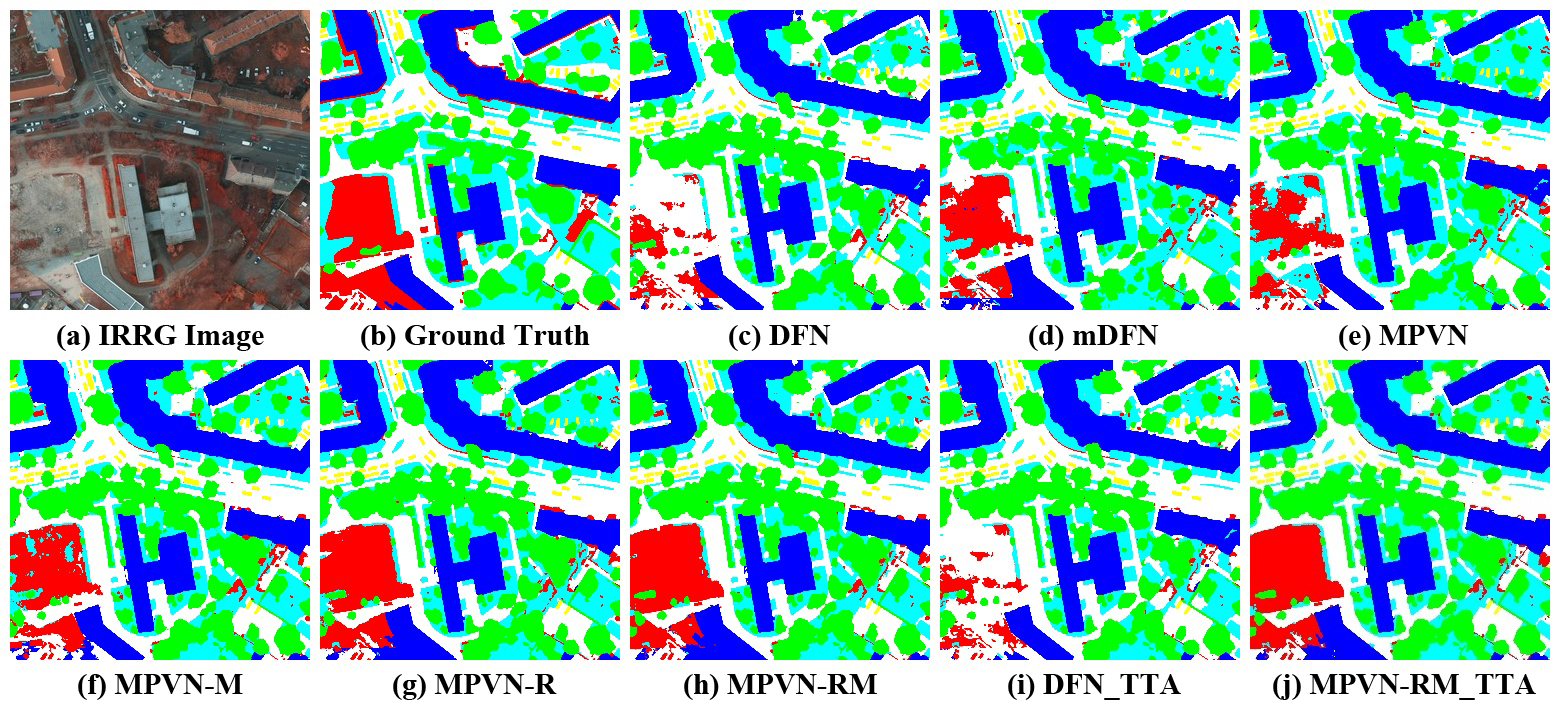}
\vfill
\includegraphics[width=0.32\linewidth]{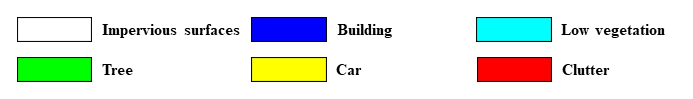}
\caption{Ablation study for our proposed AFNet on the ISPRS Potsdam 2D dataset. (a) IRRG image. (b) Ground truth. Inference result of (c) the DFN, (d) the modified DFN with stacked data (mDFN), (e) the DFN with the MPE module (MPVN), (f) the MPVN with the MAFB module (MPVN-M), (g) the MPVN with the RAFB module (MPVN-R), (h) the MPVN with the RAFB module and the MAFB module (MPVN-RM), (i) the DFN with the TTA strategy (DFN\_TTA), (j) the MPVN-RM with the TTA strategy (MPVN-RM\_TTA).}
\label{fig:ablation_study_potsdam}
\end{figure}

\begin{figure}[H]
\centering
\includegraphics[width=0.71\linewidth]{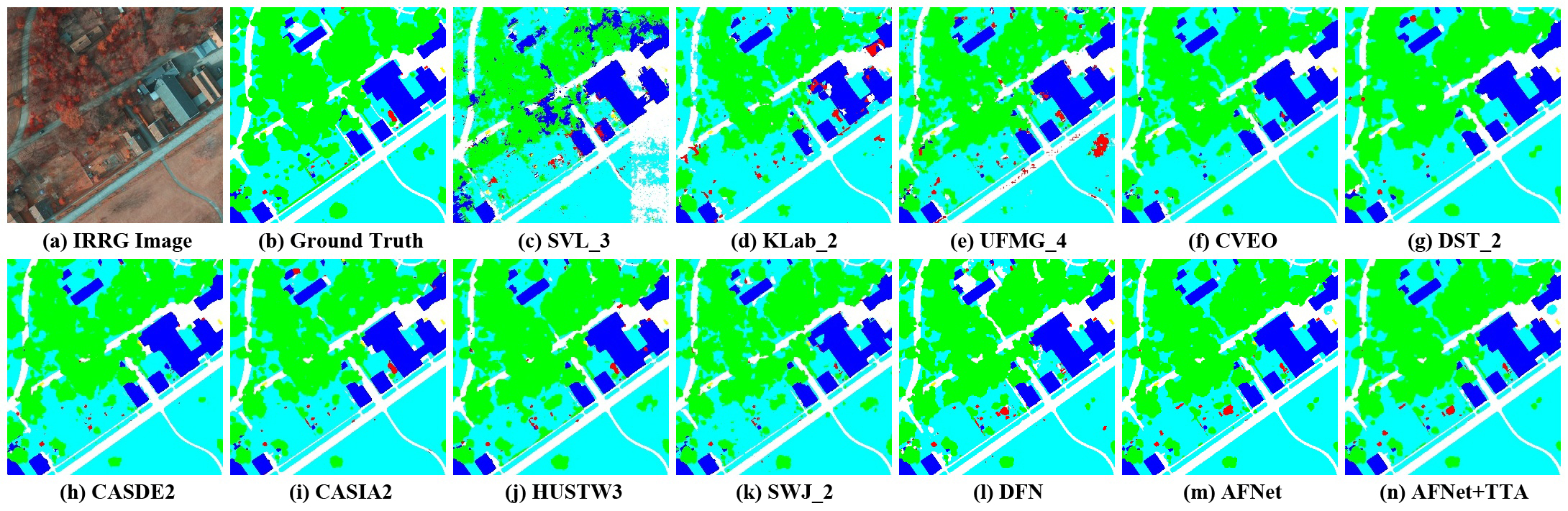}
\vfill
\includegraphics[width=0.71\linewidth]{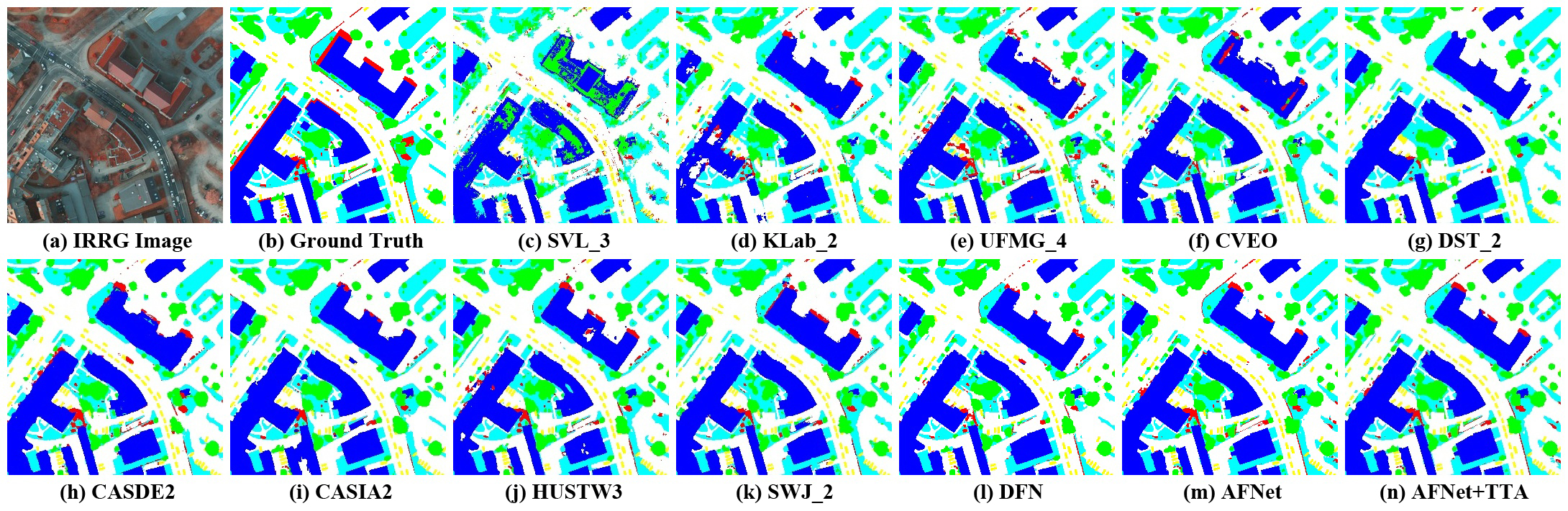}
\vfill
\includegraphics[width=0.71\linewidth]{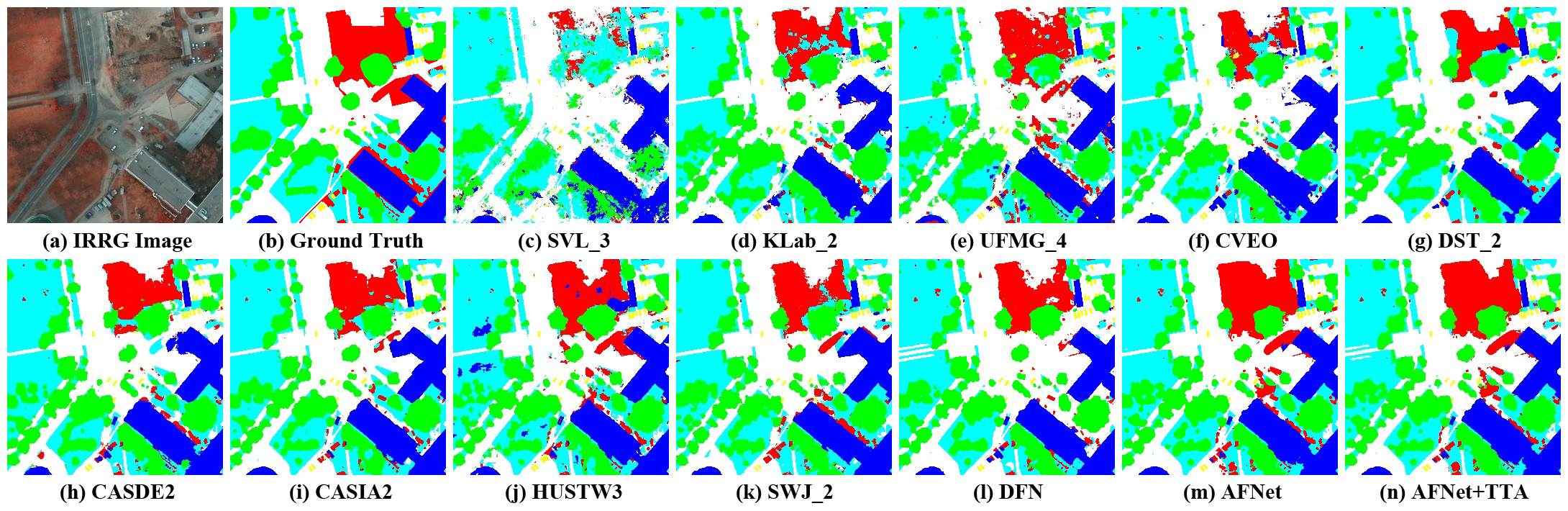}
\vfill
\includegraphics[width=0.71\linewidth]{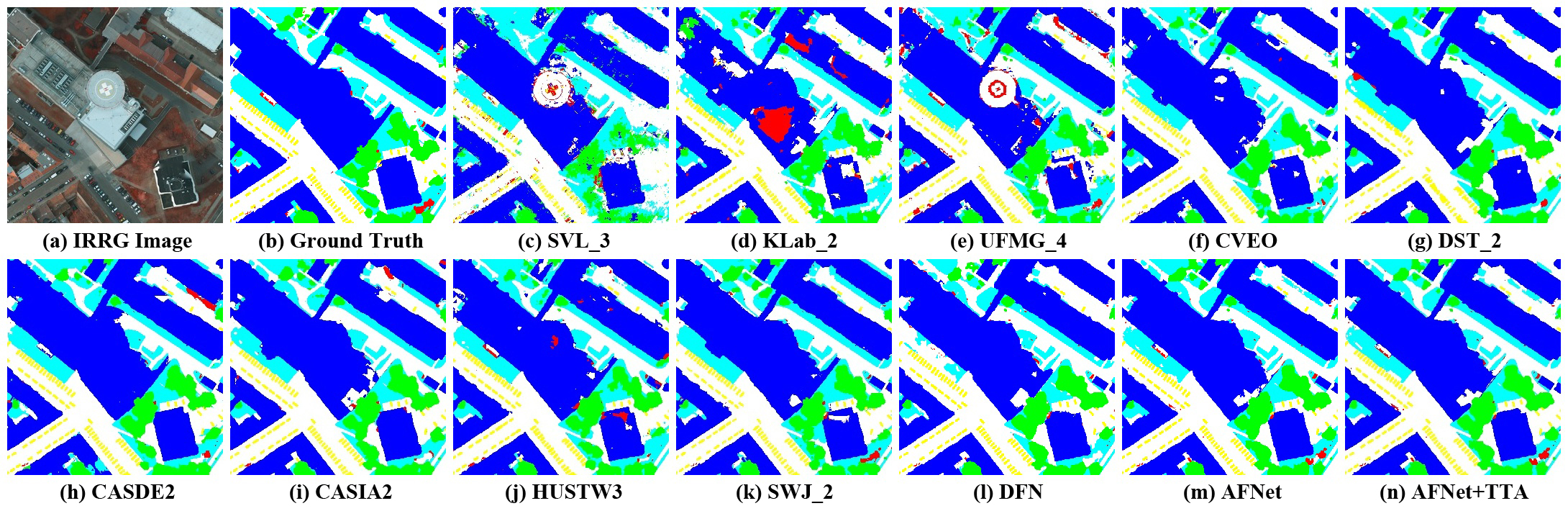}
\vfill
\includegraphics[width=0.71\linewidth]{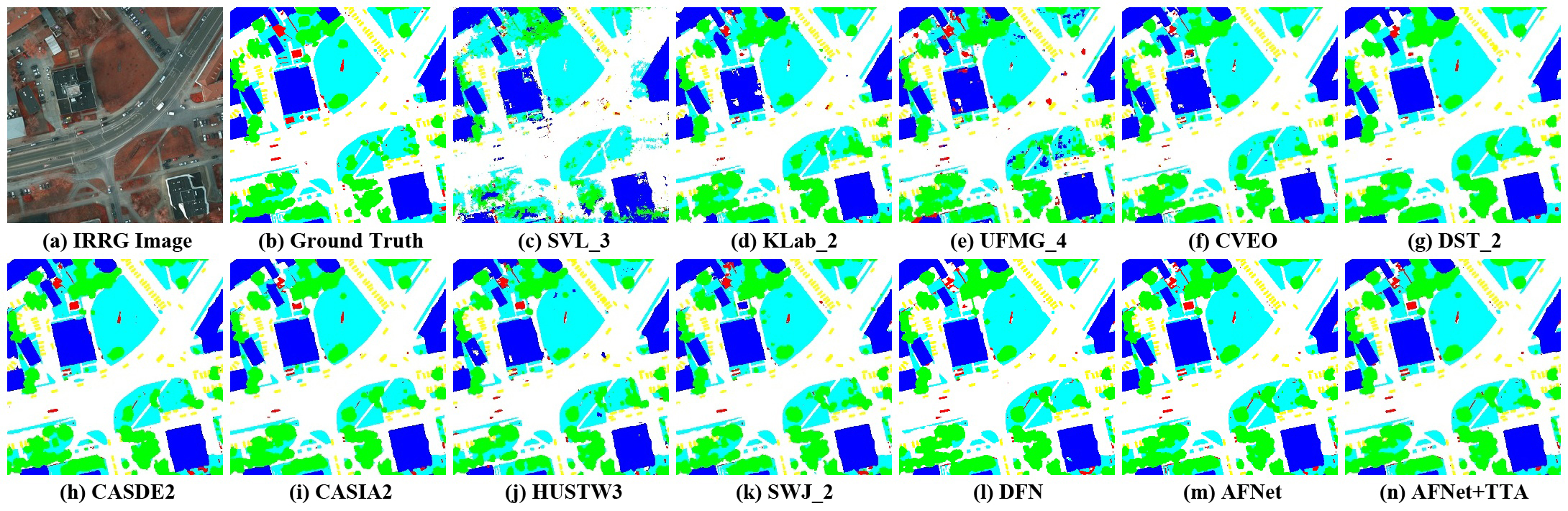}
\vfill
\includegraphics[width=0.31\linewidth]{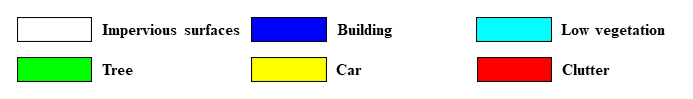}
\caption{Some examples of the results of the test set on the ISPRS Potsdam 2D dataset. Comparisons between our AFNet and other state-of-the-art methods. (a) IRRG image. (b) Ground truth. Inference result of (c) SVL\_3, (d) KLab\_2, (e) UFMG\_4, (f) CVEO, (g) DST\_2, (h) CASDE2, (i) CASIA2, (j) HUSTW3, (k) SWJ\_2, (l) the DFN, (m) our proposed AFNet, (n) our proposed AFNet with the TTA strategy.}
\label{fig:comparing_methods_potsdam}
\end{figure}

We conduct experiments on the ISPRS Potsdam 2D dataset to evaluate the effectiveness of our proposed AFNet. We apply the same training method and parameters to train the ISPRS Potsdam 2D dataset. The same inference settings are also applied to this dataset. Numerical comparisons of the ablation study for the MPE module, the RAFB module, and the MAFB module are shown in Table \ref{table:module_potsdam}. As shown in Table \ref{table:tta_potsdam}, the TTA strategy also improves the accuracy on the ISPRS Potsdam 2D dataset. Our proposed AFNet achieves a 92.1\% OA and 93.44\% mean F1 score. The detailed results of the ablation study are shown in Figure \ref{fig:ablation_study_potsdam}. Numerical comparisons with other state-of-the-art methods are shown in Table \ref{table:methods_potsdam}. According to Figure \ref{fig:comparing_methods_potsdam}, we can clearly see that our proposed AFNet outperforms all other state-of-the-art methods. Similar to the performance on the ISPRS Vaihingen 2D dataset, our AFNet has achieved state-of-the-art performance on the ISPRS Potsdam 2D dataset. Using the C++ program provided by the organizer, we also evaluated the accuracy of the AFNet inference results, and we generated a detailed evaluation webpage for the ISPRS Potsdam 2D dataset. We uploaded this evaluation webpage to our web server. The webpage is available online at \url{http://research.yangxuan.me/isprs/potsdam/radi/index.html} (accessed on 29th March 2021). 

\begin{table}[ht]
\centering
\begin{tabular}{c c c c c c c c}
\hline
\textbf{Method} & \textbf{imp\_surf} & \textbf{building} & \textbf{low\_veg} & \textbf{tree} & \textbf{car}  & \textbf{OA}   & \textbf{Mean F1} \\
\hline
DFN             & 91.0               & 97.5              & 86.1              & 89.2          & 96.4          & 90.2          & 92.04            \\
mDFN            & 93.6               & 97.1              & 86.5              & 86.1          & 96.5          & 90.6          & 91.96            \\
MPVN            & 93.1               & 97.3              & 86.6              & 88.0          & 96.7          & 90.7          & 92.34            \\
MPVN-M          & 93.2               & 97.4              & 87.8              & 89.2          & 96.6          & 91.3          & 92.84            \\
MPVN-R          & \textbf{93.9}      & \textbf{97.7}     & 88.1              & 89.0          & 96.8          & 91.7          & 93.10            \\
MPVN-RM         & \textbf{93.9}      & 97.5              & \textbf{88.4}     & \textbf{89.4} & \textbf{96.9} & \textbf{91.9} & \textbf{93.22}   \\
\hline
\end{tabular}
\caption{The effect of the MPE module, the RAFB module and the MAFB module on the ISPRS Potsdam 2D dataset.}
\label{table:module_potsdam}
\end{table}

\begin{table}[ht]
\centering
\begin{tabular}{c c c c c c c c}
\hline
\textbf{Method} & \textbf{imp\_surf} & \textbf{building} & \textbf{low\_veg} & \textbf{tree} & \textbf{car}  & \textbf{OA}   & \textbf{Mean F1} \\
\hline
DFN+TTA         & 91.2               & 97.6              & 86.4              & 89.5          & 96.6          & 90.5          & 92.26            \\
mDFN+TTA        & 93.7               & 97.2              & 86.9              & 86.2          & 96.7          & 90.8          & 92.14            \\
MPVN+TTA        & 93.4               & 97.4              & 87.1              & 88.4          & 96.9          & 91.0          & 92.64            \\
MPVN-M+TTA      & 93.3               & 97.6              & 88.2              & 89.4          & 96.8          & 91.5          & 93.06            \\
MPVN-R+TTA      & \textbf{94.1}      & \textbf{97.8}     & 88.4              & 89.2          & 97.0          & 92.0          & 93.30            \\
MPVN-RM+TTA     & \textbf{94.1}      & 97.6              & \textbf{88.7}     & \textbf{89.7} & \textbf{97.1} & \textbf{92.1} & \textbf{93.44}   \\
\hline
\end{tabular}
\caption{The effect of the TTA strategy on the ISPRS Potsdam 2D dataset.}
\label{table:tta_potsdam}
\end{table}

\begin{table}[ht]
\centering
\begin{tabular}{c c c c c c c c}
\hline
\textbf{Method} 
& \textbf{imp\_surf} & \textbf{building} & \textbf{low\_veg} & \textbf{tree} & \textbf{car}  & \textbf{OA}   & \textbf{Mean F1} \\
\hline
SVL\_3 \cite{gerke2014use} 
& 84.0               & 89.8              & 72.0              & 59.0          & 69.8          & 77.2          & 74.92            \\
KLab\_2 \cite{kemker2018algorithms} 
& 89.7               & 92.7              & 83.7              & 84.0          & 92.1          & 86.7          & 88.44            \\
UFMG\_4 \cite{nogueira2019dynamic} 
& 90.8               & 95.6              & 84.4              & 84.3          & 92.4          & 87.9          & 89.50            \\
CVEO 
& 91.2               & 94.5              & 86.4              & 87.4          & 95.4          & 89.0          & 90.98            \\
DST\_2 \cite{sherrah2016fully} 
& 91.8               & 95.9              & 86.3              & 87.7          & 89.2          & 89.7          & 90.18            \\
CASDE2 \cite{pan2018semantic} 
& 92.4               & 96.5              & 86.4              & 87.1          & 95.2          & 90.0          & 91.52            \\
DFN \cite{yu2018learning} 
& 91.0               & 97.5              & 86.1              & 89.2          & 96.4          & 90.2          & 92.04            \\
DFN+TTA 
& 91.2               & 97.6              & 86.4              & 89.5          & 96.6          & 90.5          & 92.26            \\
CASIA2 \cite{liu2018semantic} 
& 93.3               & 97.0              & 87.7              & 88.4          & 96.2          & 91.1          & 92.52            \\
HUSTW3  
& 93.8               & 96.7              & 88.0              & 89.0          & 96.0          & 91.6          & 92.70            \\
SWJ\_2 
& \textbf{94.4}      & 97.4              & 87.8              & 87.6          & 94.7          & 91.7          & 92.38            \\
AFNet (t/v) 
& 93.6               & 97.6              & 88.6              & 89.4          & 96.3          & 91.7          & 93.10            \\
AFNet (t/n) 
& 93.9               & 97.5              & 88.4              & 89.4          & 96.9          & 91.9          & 93.22            \\
AFNet+TTA (t/v) 
& 93.7               & \textbf{97.7}     & \textbf{88.8}     & 89.5          & 96.5          & 91.9          & 93.24            \\
AFNet+TTA (t/n) 
& 94.1               & 97.6              & 88.7              & \textbf{89.7} & \textbf{97.1} & \textbf{92.1} & \textbf{93.44}   \\
\hline
\end{tabular}
\caption{Accuracy comparisons between our AFNet and other state-of-the-art methods on the ISPRS Potsdam 2D dataset.}
\label{table:methods_potsdam}
\end{table}

\section{Discussion}
\label{sec:5}

\subsection{Encoder}

The MPE module used by AFNet in this paper includes two branches, ResNet-50 and ResNet-18. The main branch uses ResNet-50. Because there are only 16 tiles of images in the training samples of the ISPRS Vaihingen 2D dataset, an overly large encoder is not needed. 

\begin{figure}[ht]
\centering
\subfigure[]{
    \includegraphics[width=0.45\linewidth]{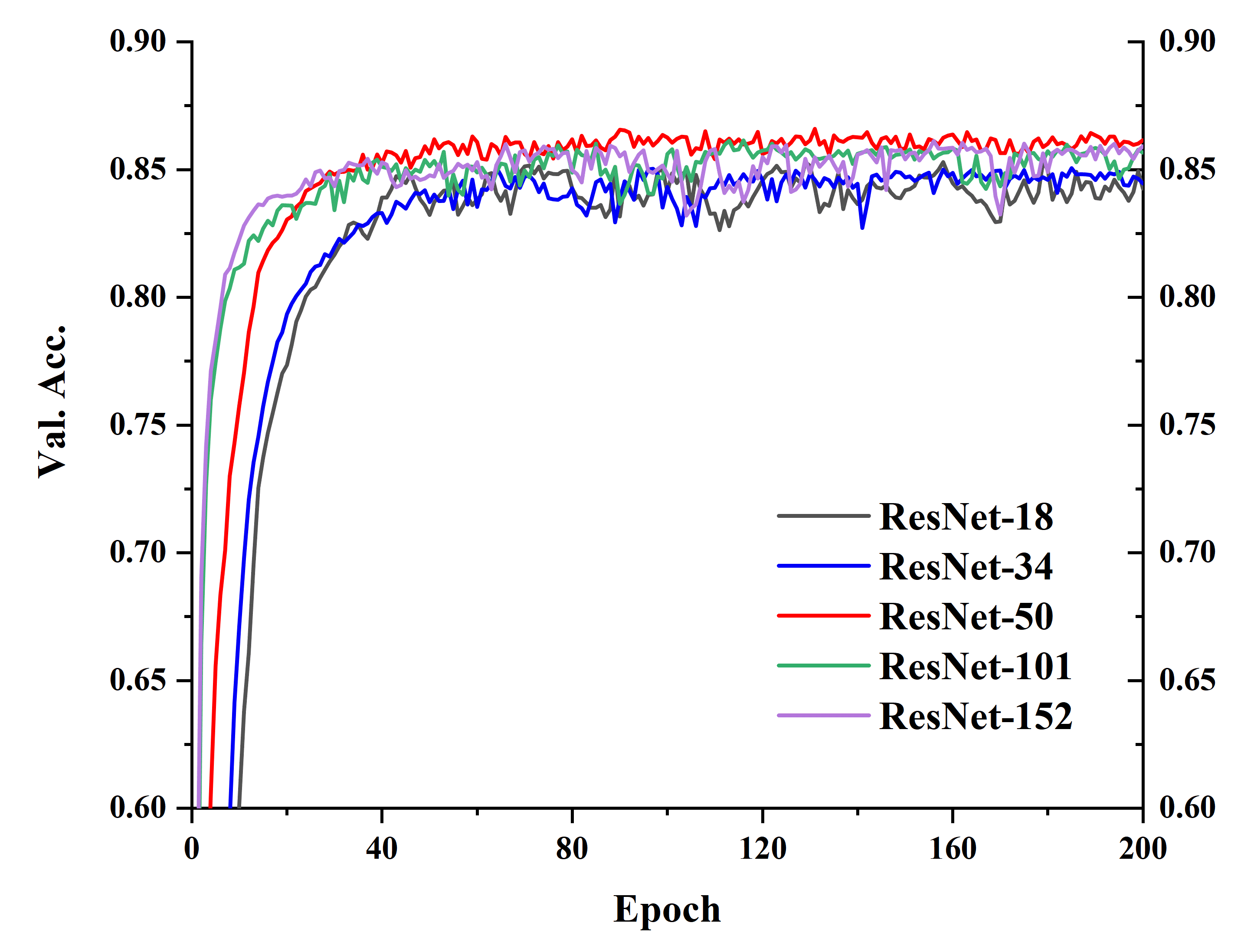}
    \label{fig:encoder:a}
}
\subfigure[]{
    \includegraphics[width=0.45\linewidth]{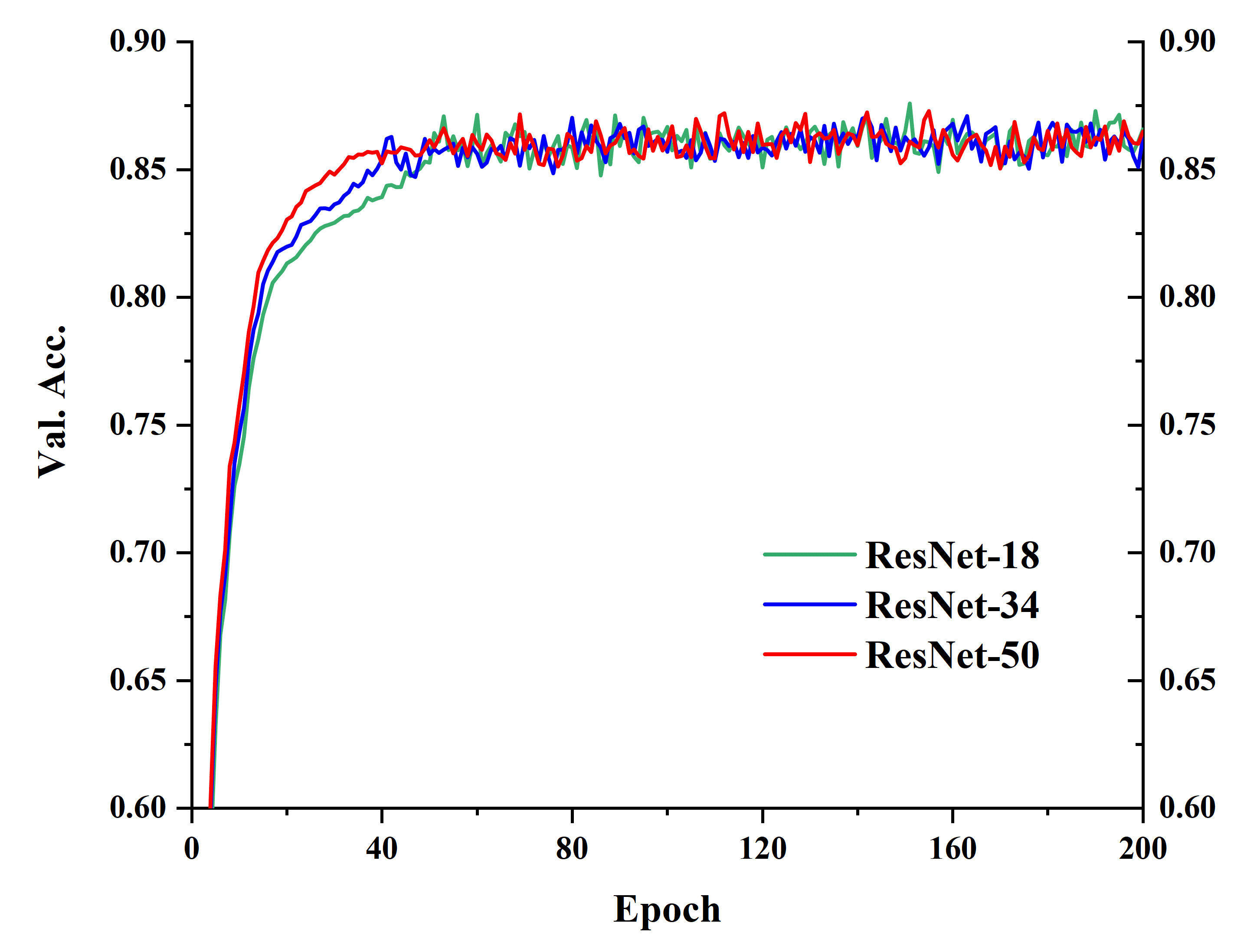}
    \label{fig:encoder:b}
}
\caption{The accuracy of different encoders on the validation set. (a) Main branch of the MPE. (b) Auxiliary branch of the MPE.}
\label{fig:encoder}
\end{figure}

The most common ResNet has five types, including ResNet-18/34/50/101/152. First, we choose ResNet-50 as the baseline of the main branch of the encoder. Then, we test a replacement of ResNet-50 with ResNet-18/34. During the training, we find that the accuracy of the validation set cannot reach the baseline accuracy, which is approximately 1\% lower, as shown in Figure \ref{fig:encoder:a}. The reason is that the feature abstraction ability of ResNet-18/34 is weak and cannot meet the complexity requirement of the dataset. Next, we test a replacement of ResNet-50 with ResNet-101/152. During the training, we find that the accuracy of the validation set also cannot reach the baseline accuracy, which is approximately 0.5\% lower, as shown in Figure \ref{fig:encoder:a}. The reason is that ResNet-101/152 has too many layers, but the dataset is relatively small. Therefore, the performance of the network deteriorates from the baseline. Therefore, we ultimately adopt ResNet-50 as the main branch of the encoder. 

Next, we discuss the auxiliary branch of the encoders. First, we choose ResNet-18 as the baseline of the auxiliary branch of the encoder. Because NDVI/DSM data can be regarded as a simple low-level feature after simple coding, the feature complexity is relatively low. Therefore, it does not require a deep network as the auxiliary branch of the encoder. Another reason is that using ResNet-50 as both the main branch and the auxiliary branch increases the total number of network parameters. However, GPU resources are limited and cannot handle the amount of these parameters. We set the slice size to 512 to compare the performance of different auxiliary branches of the encoder. We find that most abstract features are extracted from IRRG data in the experiment. NDVI/DSM data are only supplementary for limited improvement. As shown in Figure \ref{fig:encoder:b}, ResNet-18, ResNet-34, and ResNet-50 have almost the same performance. Therefore, we ultimately adopt ResNet-18 as the auxiliary branch of the encoder. 

\begin{figure}[ht]
\centering
\includegraphics[width=0.45\linewidth]{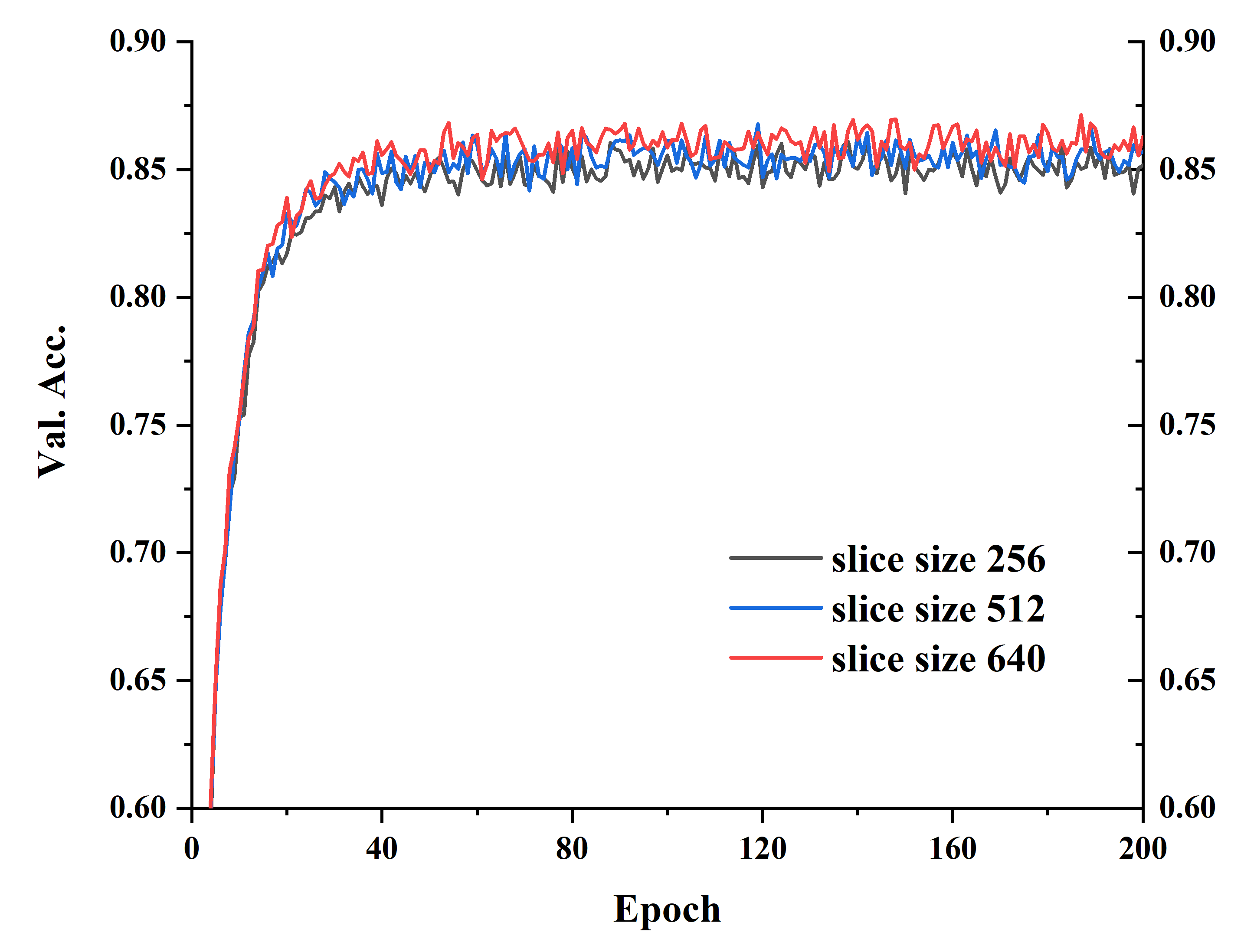}
\caption{The accuracy of different slice sizes on the validation set.}
\label{fig:slice}
\end{figure}

\subsection{Slice Size}

In this paper, we set the slice size of the network input to 640. In the experiment, we find that the larger the slice size in the training stage is, the better the training performance. We set the slice size to 256, 512, and 640. As shown in Figure \ref{fig:slice}, when the slice size is set to 640, the accuracy of the validation set is the highest. The most likely reason is that the remote sensing scene is very complicated, and the scale span of different target objects is very large. There may only be one category of the target object in a slice, which makes the network training unstable. To avoid this problem, we need to set the slice size as large as possible. Due to the GPU memory limitation, the larger the slice size is, the smaller the batch size. However, because the batch normalization (BN) \cite{ioffe2015batch} layer is used in AFNet, the batch size should not be too small, which is a trade-off problem. The BN layer normalizes the data of one batch, so the BN layer is very sensitive to the distribution of the element values of all the image data or the feature map. A batch size that is too small will cause the mean value and standard deviation in the BN layer to be unstable, which will have a negative impact on the network training. As shown in Figure \ref{fig:ablation_study}, the shapes of the target objects in the same category are very similar, and the gray histogram distribution of these images are almost the same. Therefore, on this dataset, the effect of batch size on BN layers is negligible. To keep the BN layer in ResNet to load the pre-trained model, we set the batch size to 2. The maximum input size of AFNet is 640 for the NVIDIA TITAN Xp GPU. 

In addition, since we use the random crop data augmentation strategy, the slice size should be larger than the input size of the network. To make the random crop strategy as random as possible, we set the slice size to 800, which allows 25,600 possibilities for random cropping per slice, and all randomly cropped slices can be fed into the network. This setting greatly increases the dataset. 

\begin{figure}[ht]
\centering
\includegraphics[width=0.45\linewidth]{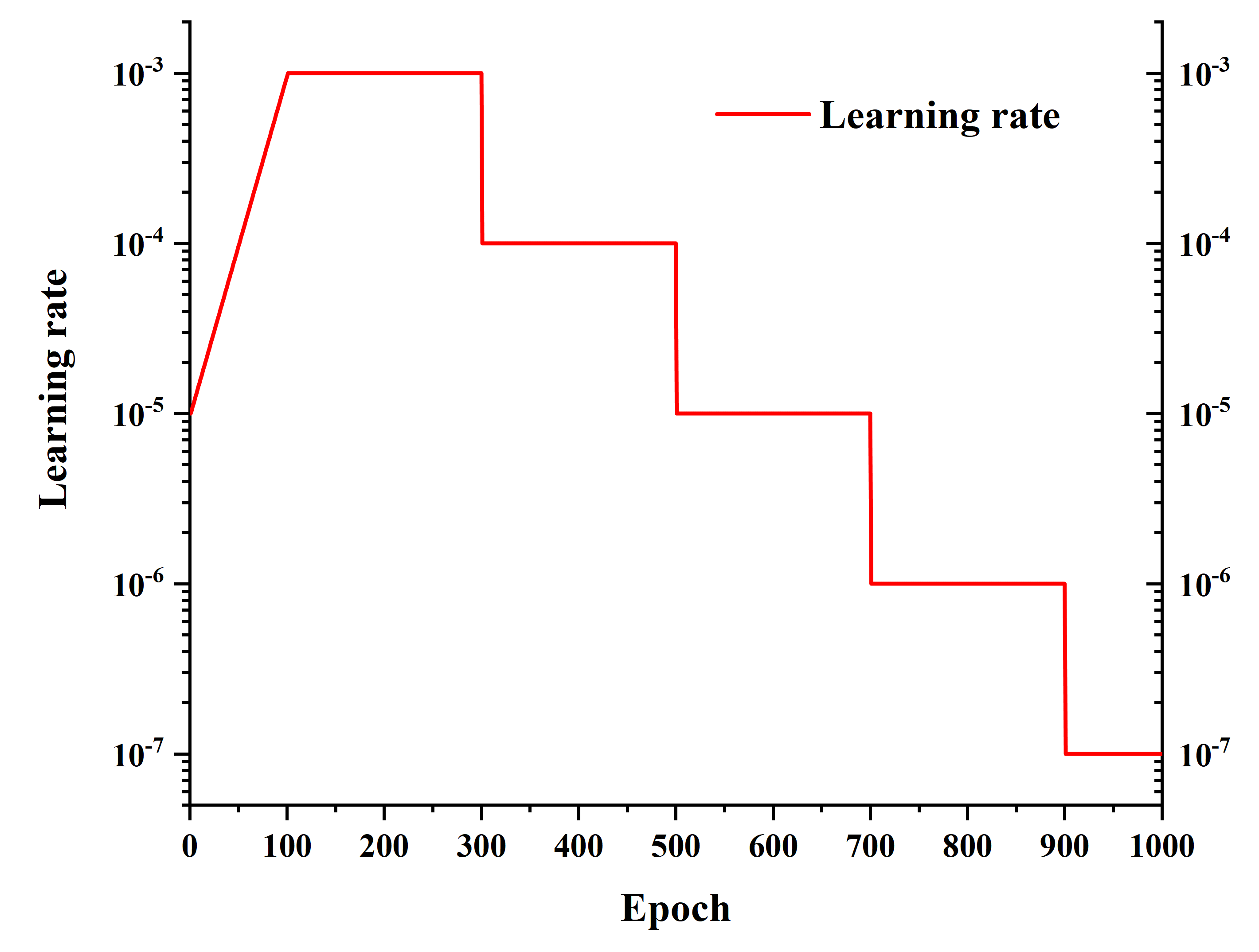}
\caption{The learning rate curve with the WarmUp strategy and the Step strategy.}
\label{fig:optim_lr}
\end{figure}

\subsection{Optimization}

The learning rate strategy in this paper uses the WarmUp strategy and the Step strategy. The MPE module of the AFNet contains ResNet-50 and ResNet-18. Both of these encoders have pre-trained models on the ImageNet dataset, which can speed up the convergence of the network if using the pre-trained model to initialize the MPE module parameters in the AFNet. Additionally, the model accuracy is improved to some extent. The ImageNet dataset contains millions of samples. At present, remote sensing datasets are far from reaching the order of magnitude of ImageNet. The labeling quality is not as good as ImageNet. Therefore, we use the pre-trained model on the ImageNet dataset to initialize the MPE module parameters. However, the ImageNet dataset contains natural images, and the imaging content, angle, and gray histogram distribution are significantly different from those of remote sensing images. The pre-trained model parameters on the ImageNet dataset are very different from those when the network finally converges. When the learning rate is set low, such as $1\times 10^{-5}$, the MPE module parameters tend to fall into a local minimum that is close to the distribution of the ImageNet dataset, resulting in the whole network being unable to converge to a globally optimal solution. When the learning rate is set high, such as $1\times 10^{-3}$, the gradient update of the network is too fierce, which causes severe jittering in the MPE module parameters in the network, and the meaning of initializing the MPE module parameters from the pre-trained model is lost. Therefore, we adopt the WarmUp strategy and set the initial learning rate to $1\times 10^{-5}$. With an increasing number of iterations, the learning rate gradually increases to $1\times 10^{-3}$, which can ensure that the MPE module parameters gradually adapt to the distribution of remote sensing data from the distribution of the ImageNet dataset. Simultaneously, a high learning rate can ensure that the parameters of the whole AFNet have sufficient learning motivation. In the middle and late periods, the network parameters are close to convergence. If the learning rate is too large, the parameters will oscillate and fail to converge to the global minimum. Therefore, the Step strategy is adopted in this paper. The learning rate is multiplied by a factor of 0.1 every 200 epochs. We find that the network convergence is more stable, and the network performance is slightly improved after using the learning rate Step strategy in our experiments. The learning rate curve with the WarmUp strategy and the Step strategy is shown in Figure \ref{fig:optim_lr}. 

\begin{figure}[ht]
\centering
\subfigure[]{
    \includegraphics[width=0.45\linewidth]{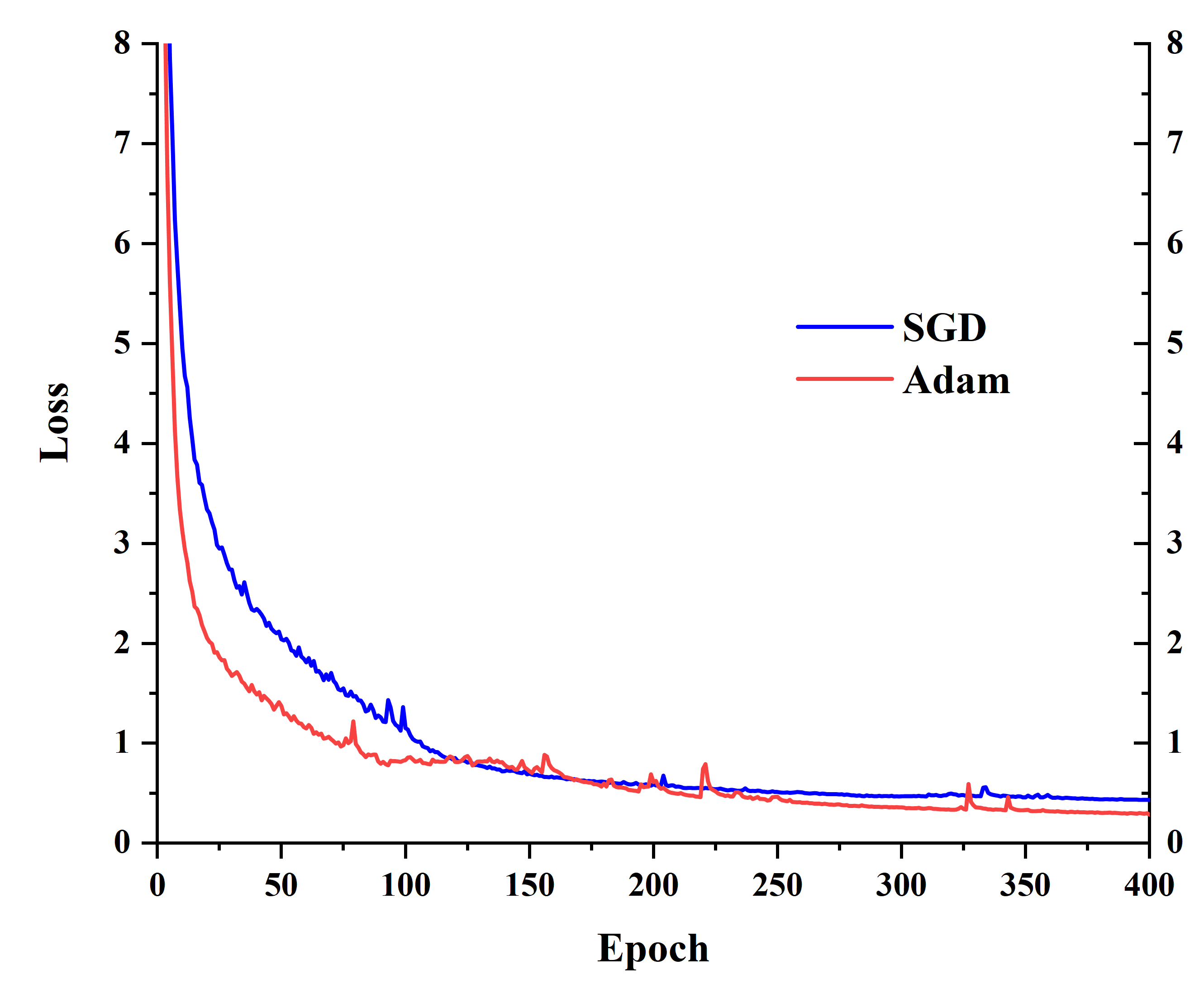}
    \label{fig:optim_optim:a}
}
\subfigure[]{
    \includegraphics[width=0.45\linewidth]{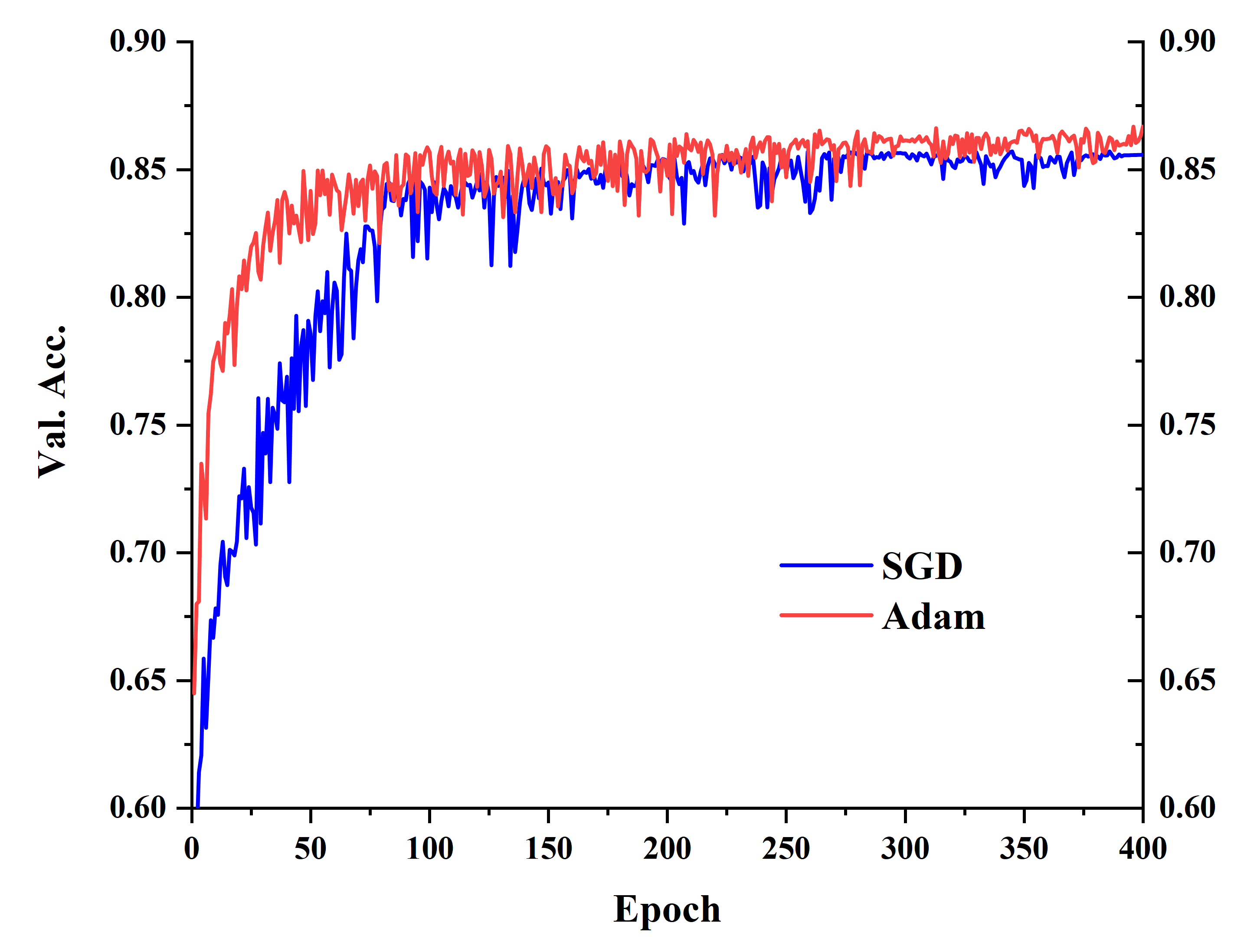}
    \label{fig:optim_optim:b}
}
\caption{The difference between the SGD optimizer and Adam optimizer. (a) The training loss value curve. (b) The accuracy on the validation set.}
\label{fig:optim_optim}
\end{figure}

We choose adaptive moment estimation (Adam) \cite{kingma2014adam} as the AFNet optimizer. Adam is a first-order optimization algorithm that can replace the traditional stochastic gradient descent (SGD) process. The optimizer can iteratively update the parameters of the network based on training samples. The SGD optimization algorithm is sensitive to the learning rate. To take advantage of the SGD optimization algorithm, a precise strategy is necessary for the learning rate. The Adam optimization algorithm is not sensitive to the learning rate. Adam dynamically adjusts the learning rate within a certain range according to the gradient update amplitude. Therefore, Adam can make the network converge quickly. Our proposed AFNet contains the MPE module with two encoder branches and the more complicated MAFB module and RAFB module in the decoder, which makes it difficult for the network parameters to converge. It is difficult to design a valid learning rate strategy with the SGD optimization algorithm. As shown in Figure \ref{fig:optim_optim}, we find that the parameters of the network converge slowly if we use SGD as the AFNet optimizer in the experiments. The loss value with SGD is significantly greater than the loss value with Adam when the network parameters converge. The performance of the network trained by SGD is not as good as that of the network trained by Adam. In contrast, Adam can accelerate the convergence of AFNet and does not require a complex learning rate strategy. Therefore, we choose Adam as the AFNet optimizer, and achieve the best performance on the ISPRS Vaihingen 2D dataset. 

\subsection{Overfitting}

In this paper, the samples are divided into a training set and a validation set according to different proportions, which are adjusted at two different stages. In the stage of network design and debugging, two samples are selected as the validation set, which is used to evaluate the network performance in real-time during the training and stop the training in time to avoid overfitting. At this stage, different learning rate strategies, optimizers, and loss functions are used to determine a better training setting. 

\begin{figure}[ht]
\centering
\subfigure[]{
    \includegraphics[width=0.45\linewidth]{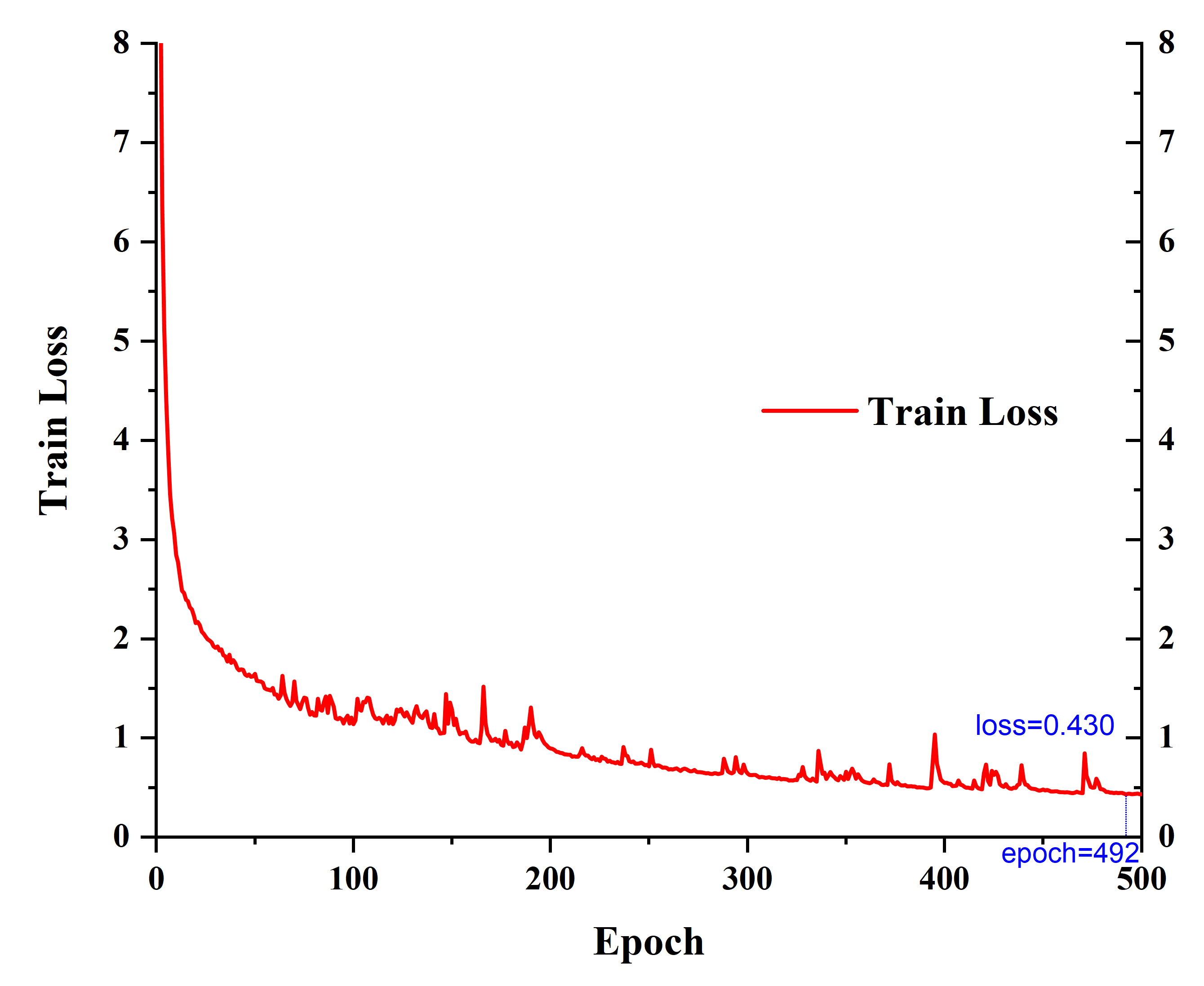}
    \label{fig:overfitting:a}
}
\subfigure[]{
    \includegraphics[width=0.45\linewidth]{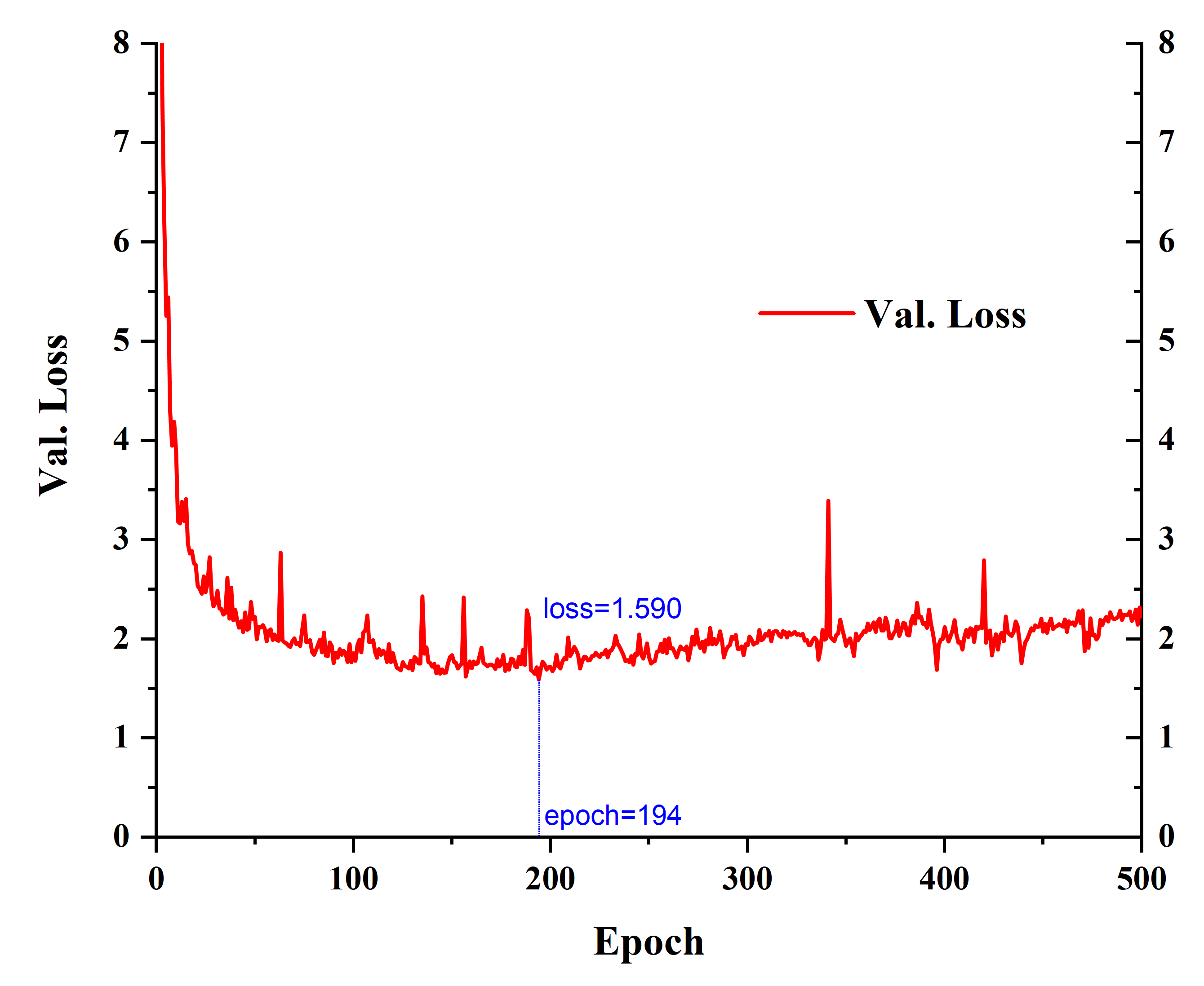}
    \label{fig:overfitting:b}
}
\vfill
\subfigure[]{
    \includegraphics[width=0.45\linewidth]{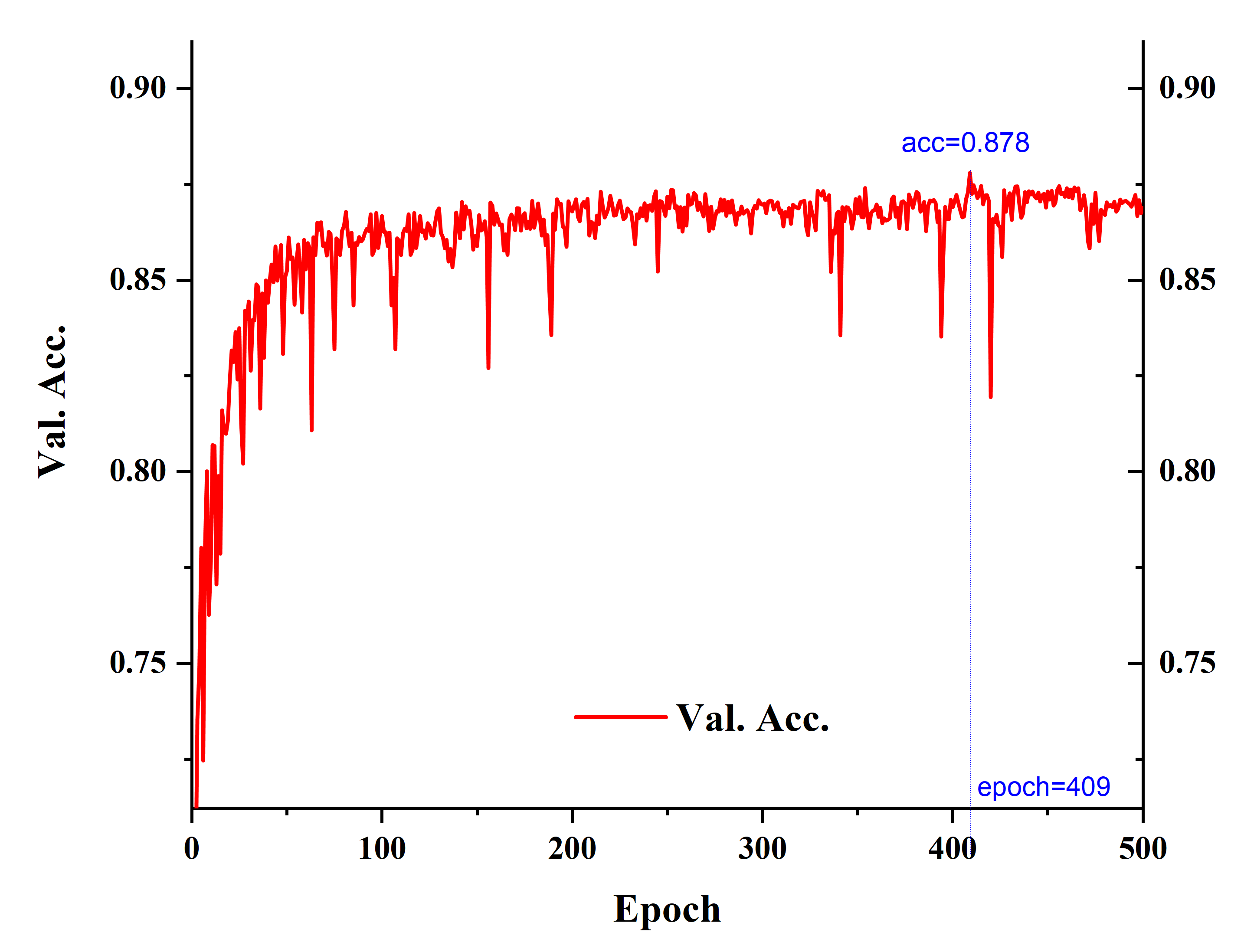}
    \label{fig:overfitting:c}
}
\caption{Training curves about training loss, validation loss, and validation accuracy on the ISPRS Vaihingen 2D dataset. (a) The training loss value curve. The minimum loss is 0.430 when epoch is 492. (b) The validation loss value curve. The minimum loss is 1.590 when epoch is 194. (c) The accuracy on the validation set. The maximum accuracy is 0.878 when epoch is 409.}
\label{fig:overfitting}
\end{figure}

As shown in Figure \ref{fig:overfitting}, we find that the training loss continues to maintain a downward trend. The validation loss starts to increase at the 194th epoch, which means that the model begins to overfit. However, the validation accuracy has a slight upward trend in general and reaches the best accuracy at the 409th epoch. The final output of the network is a probability feature map, which is thresholded to obtain the predicted result. Although the reliability of the model on the validation set has decreased slightly, it can still exceed the threshold for correct classification. Therefore, although the model has a slight tendency to overfit, the validation accuracy is not affected by the overfitting. 

We take advantage of the feature that slight overfitting will not affect the validation accuracy and use the two images to tune the network. We fix the number of training epochs and choose the best training settings according to the validation accuracy. To make full use of all 16 tiles of trainable samples, we combine the training set and validation set and retrain the model using all available data. Since there is no independent validation set, we set the learning rate strategy, optimizer, and loss function to be the same as previous settings. We use the same number of training epochs that align with the previous approach. The loss values and the accuracy values in several key epochs on the ISPRS Vaihingen 2D dataset are shown in Table \ref{table:overfitting}. We obtain the best inference results that achieve the best performance on the ISPRS Vaihingen 2D dataset. 

Since overfitting should be avoided for most datasets, we perform the same experiment on the ISPRS Potsdam 2D dataset to confirm whether this method is applicable. As shown in Table \ref{table:overfitting_potsdam}, we find a similar pattern to the ISPRS Vaihingen 2D dataset. Therefore, we first train a model to find the number of training epochs and best training setting. Then, we use all 24 tiles of trainable samples to retrain the network and use the same number of epochs that align with the previous approach. This model achieves the best performance on the ISPRS Potsdam 2D dataset. However, it should be noted that overfitting is still not recommended on most datasets and may only be used on these two datasets. 

\begin{table}[ht]
\centering
\begin{tabular}{c c c c c}
\hline
\textbf{Epoch} & \textbf{Train Loss} & \textbf{Val Loss} & \textbf{Val Accuracy} & \textbf{Test Accuracy} \\
\hline
194            & 1.057               & 1.590             & 87.0                  & 90.8                   \\
409            & 0.523               & 1.889             & 87.8                  & 91.7                   \\
492            & 0.430               & 2.241             & 87.0                  & 91.1                   \\
\hline
\end{tabular}
\caption{The loss values and the accuracy values in several key epochs on the ISPRS Vaihingen 2D dataset. The training loss value is minimum when the epoch is 492. The validation loss value is minimum when the epoch is 194. The validation accuracy value is maximum when the epoch is 409.}
\label{table:overfitting}
\end{table}

\begin{table}[ht]
\centering
\begin{tabular}{c c c c c}
\hline
\textbf{Epoch} & \textbf{Train Loss} & \textbf{Val Loss} & \textbf{Val Accuracy} & \textbf{Test Accuracy} \\
\hline
35             & 2.000               & 2.507             & 89.0                  & 91.9                   \\
46             & 1.802               & 2.578             & 89.1                  & 92.1                   \\
80             & 1.632               & 2.671             & 88.9                  & 91.9                   \\
\hline
\end{tabular}
\caption{The loss values and the accuracy values in several key epochs on the ISPRS Potsdam 2D dataset. The training loss value is minimum when the epoch is 80. The validation loss value is minimum when the epoch is 35. The validation accuracy value is maximum when the epoch is 46.}
\label{table:overfitting_potsdam}
\end{table}

\section{Conclusions}
\label{sec:6}

In this paper, we proposed a new method for semantic segmentation of very-high-resolution remote sensing imagery. We designed the MPE structure to extract the IRRG image feature and the NDVI/DSM auxiliary feature. The two branches of the MPE are asymmetric, which can extract different types of features from different data according to different characteristics, simultaneously saving hardware resources and ensuring accuracy. Based on the DFN and MPE, we proposed the MPVN. Inspired by the CA structure and SA structure, which allow the network to learn the effective information of channel dimensions and spatial dimensions by itself, we designed the MAFB module and RAFB module. The MAFB module can efficiently fuse the different types of features and allows the network to learn the effective information in the different types of data by itself. The RAFB module can efficiently fuse the high-level abstract features and low-level spatial features. Based on our proposed MPVN with the MPE, MAFB, and RAFB, we proposed the AFNet to solve the data fusion and data mining problem in very-high-resolution remote sensing imagery. We experimented with our proposed AFNet on both the ISPRS Vaihingen 2D dataset and the ISPRS Potsdam 2D dataset and achieved state-of-the-art performance compared with other methods. In future research, we will promote our proposed AFNet to more datasets. 

\section*{Acknowledgments}

The authors thank the International Society for Photogrammetry and Remote Sensing (ISPRS) for making the Vaihingen dataset and the Potsdam dataset available online.

The authors thank the editors and anonymous reviewers for their valuable comments, which greatly improved the quality of the paper.

This research was supported by the Strategic Priority Research Program of the Chinese Academy of Sciences under Grant No. XDA19080302. 


\bibliography{yang-afnet-isprs-bib}

\end{document}